\documentclass[11pt]{article}

\usepackage{amsmath}
\usepackage[preprint]{acl}

\usepackage{times}
\usepackage{latexsym}

\usepackage[T1]{fontenc}

\usepackage[utf8]{inputenc}

\usepackage{microtype}

\usepackage{inconsolata}

\usepackage{graphicx}
\usepackage{fancyhdr}
\usepackage{setspace}
\usepackage{cleveref}
\usepackage{tikz}

\usepackage{amssymb}
\usepackage{booktabs}
\usepackage{tipa}
\usepackage{subcaption}
\usepackage{makecell}
\usepackage{soul}
\usepackage{gb4e}
\noautomath

\newcommand\autocite{\citep} \newcommand\textcite{\citet}

\title{Beyond Decodability: Reconstructing Language Model Representations with an Encoding Probe}

\author{ Gaofei Shen\textsuperscript{1} ~ Martijn Bentum\textsuperscript{2}  ~
      Tomas O. Lentz\textsuperscript{1}  ~ Afra Alishahi\textsuperscript{1}  ~
      Grzegorz Chrupa\l{}a\textsuperscript{1}
    \\
    \textsuperscript{1}Tilburg University, \textsuperscript{2}Radboud University
    \\
    \texttt{\{g.shen, a.alishahi, t.o.lentz\}@tilburguniversity.edu}\\
    \texttt{martijn.bentum@ru.nl}, \texttt{grzegorz@chrupala.me}\\
    }

\begin{document}
\maketitle

\begin{abstract}
  Probing is widely used to study which features can be decoded from language
  model representations. However, the common decoding probe approach has two
  limitations that we aim to solve with our new encoding probe approach:
  contributions of different features to model representations cannot be
  directly compared, and feature correlations can affect probing results. We
  present an \textsc{Encoding Probe} that reverses this direction and
  reconstructs internal representations of models using interpretable features.
  We evaluate this method on text and speech transformer models, using feature
  sets spanning acoustics, phonetics, syntax, lexicon, and speaker identity. Our
  results suggest that speaker-related effects vary strongly across different
  training objectives and datasets, while syntactic and lexical features
  contribute independently to reconstruction. These results show that the
  \textsc{Encoding Probe} provides a complementary perspective on interpreting
  model representations beyond decodability.
\end{abstract}

\section{Introduction}
\label{sec:intro}
Most current approaches to processing speech and text are based on deep learning
architectures and especially large transformer models. While these systems work
very well in many applications, they are large and complicated enough that
understanding and controlling their function in detail becomes challenging. A
standard approach to analyzing deep learning models is based on \textit{probing}
the representations which emerge in these systems. More specifically, a
researcher would come up with a set of hypotheses about what kinds of
information a model can be plausibly expected to encode and then attempt to
systematically extract that information from activation patterns in different
components of the system.

\subsection{Decoding probe and its limitations}
\label{sec:decoding-probe-limitations}
This classic probing approach, which we refer to in the rest of the paper as
\textsc{Decoding Probe}, can tell us whether a specific information type can be
reliably extracted or \textit{decoded} from a model's representations. This
approach offers valuable insights, and can be indispensable, for example when we
want to know whether a model stores sensitive information such as speaker
identity that has privacy implications. However, this approach has two important
limitations that we aim to solve with our new \textsc{Encoding Probe} approach.

\paragraph{Relative feature contributions.}
Suppose we are interested in the encoding of speaker identity versus phonetics
in the activation patterns of a model of spoken language. We train two
\textsc{Decoding Probes}, one mapping activation patterns to speaker IDs and the
other one mapping those same activation patterns to phone labels. In our
experiments on wav2vec2-base, a decoding probe recovers speaker IDs with
$94.5\%$ balanced accuracy and phone labels with $70.3\%$ balanced accuracy at
the respective best-performing layer (for the complete decoding probe metrics,
see \Cref{tab:decoding-results}). What can we conclude about the relative
importance of these two features? Nothing definitive: while we can normalize the
scores based on the number of labels and their distributions, the two numbers
are not directly comparable. These scores do not live on a common scale of
\emph{contribution} to the representation. A high decoding score tells us only
that information is available, not how much it shapes the representation.

\paragraph{Feature correlations.}
Consider a further problematic scenario: we are interested in the encoding of
grammatical categories in a language model. We train a probing classifier which
maps activation patterns to the grammatical category label for the current
token. The classifier performs at 85\% accuracy and we conclude that grammatical
category is encoded in model representations.  However, this conclusion may be
misleading if we do not \textit{control for correlated features}. Word identity
and grammatical category are highly correlated: many words occur almost
exclusively with a single grammatical category. As a result, the probe may rely
primarily on information about word identity present in the representations, and
predict grammatical categories as a by-product of their correlation with
specific words. The probe's performance may reflect the model's encoding of
lexical identity rather than grammatical category itself.

\subsection{Reversing the direction of probing}
Within NLP, the terms \textit{probe} and \textit{probing classifier} have often
been used in a narrow sense,
as a classifier which maps an intermediate representation $X$ from a target
model to some (linguistic) property of interest $Y$
\autocite[e.g.,][]{belinkovProbingClassifiersPromises2022}.
However, two assumptions in this definition can (and should) be dropped. First,
the mapping need not take the form of a classifier since the targeted property
need not be categorical. More fundamentally, the mapping need not be in the
direction from model representation $X$ to property of interest $Y$, $f: X \to
  Y$, but could be in the reverse direction $g: Y \to X$. Within neuroscience this
reverse direction of probing, known as brain encoding, is widely used and well
understood \autocite{ivanova2021probing}.
Here we call the type of probe formulated as $f: X \to Y$ a \textsc{Decoding
  Probe}, and the one formulated as $g: Y \to X$ an \textsc{Encoding Probe}.

We argue that the two directions of probing answer complementary sets of
questions about the nature of neural representations, and they have quite
different affordances in terms of how they can be implemented and applied.
Below, we discuss the advantages of the \textsc{Encoding Probe} for addressing the
two main challenges highlighted before.

\subsubsection{Quantifying feature contributions}
Consider again comparing the contribution of two separate features to the
representations of a spoken language model. We can operationalize this
comparison in terms of explaining the amount of variance in these
representations. Using an \textsc{Encoding Probe}, we can fit a regression model
to reconstruct activation patterns $X$ based on a number of interpretable
features $Y$, which include both speaker IDs $Y_{sid}$ and phone labels
$Y_{ph}$, formulated as $g: Y \to X$. We can now ablate the full probe, removing
each feature of interest in turn: first, $g_1: Y \setminus Y_{sid} \to X$, and
second, $g_2: Y \setminus Y_{ph} \to X$. Let $E(g)$ be the measure of
reconstruction error for probe $g$. Then the increase in error due to ablating
each feature, $E(g_i) - E(g)$, gives us a direct measure of the relative
contribution of each feature to the model representation, over and above the
other features. We may well find that the contribution of phone labels is higher
than that of speaker labels, even if the latter are easier to decode.

\subsubsection{Controlling for correlated features}
Reconsider the question of whether grammatical categories are encoded in a
language model. By employing an \textsc{Encoding Probe}, we can control for
correlated features: our full model $g: Y \to X$ contains multiple, potentially
correlated, sources of information, including word identities $Y_{w}$ and
grammatical category labels $Y_{g}$. We can ablate each feature separately, as
well as both together, and measure the resulting change in reconstruction error.
This allows us to assess the conditional contribution of each feature, while
accounting for their correlation.
We may find that removing $Y_{g}$ does not increase reconstruction error,
whereas removing $Y_{w}$ does. This would indicate that $Y_{g}$ does not provide
additional information beyond what is already captured by $Y_w$, even though
grammatical labels may be easy to recover from the model's representations due
to their correlation with word identity. Conversely, if ablating $Y_g$ does
increase reconstruction error, this indicates that grammatical category
contributes information to the representations that is not fully explained by
word identity.

No doubt many readers recognize that in the above vignettes we are describing
standard tools and insights of regression analysis, applied to the specific case
of analyzing model representations. This connection to well-understood
statistical methods is indeed one of the main attractions of the
\textsc{Encoding Probe} formulation.

In the remainder of the paper we apply the \textsc{Encoding Probe} to a set of
speech and text transformer models with controlled feature ablations. We focus
on syntactic and lexical features in both modalities, and on speaker identity,
phonetic, and acoustic features in speech models. Overall, the results
demonstrate the value of the \textsc{Encoding Probe} for measuring relative
contributions while controlling for correlated features.

\section{Literature}
\label{sec:literature}

Probing has been a cornerstone in the interpretability literature since the early
days of its development
\autocite{alainUnderstandingIntermediateLayers2017,conneauWhatYouCan2018,hupkesVisualisationDiagnosticClassifiers2018}.
In transformer interpretability literature, the probing paradigm has been used
to discover a variety of features encoded in the hidden state activations of
transformer models. However, \textcite{belinkovProbingClassifiersPromises2022}
identifies several important caveats in probing: probe scores are often hard to
interpret and compare across/within tasks; results can reflect probe design
choices rather than properties of the target model; probes establish correlation
rather than causation; outcomes are sensitive to dataset/task artefacts; and
analyses are constrained by pre-selected properties.

To improve interpretability of probe results,
\textcite{hewittDesigningInterpretingProbes2019} argue that a control task is
needed to ascertain that a successful probe is uncovering linguistic structure
from model internals rather than learning the task itself. In follow-up work,
\textcite{hewittConditionalProbingMeasuring2021} introduce \emph{conditional
  probing} in order to control for correlated features. Their approach relies on
the decoding probe, but compares decoding a property (such as POS label) from
target representation only, versus decoding it from the representation paired
with a control variable (such as word identity).

Our \textsc{Encoding Probe}, in contrast, controls for such effects
simultaneously for multiple properties, while also quantifying their relative
contribution to explaining the overall variance of neural representations. Our
method does not address other caveats associated with probing. While enabling
control for correlated features, the \textsc{Encoding Probe} remains
observational and not interventional. Hence it cannot establish causal structure
among the investigated variables or their role in the computations carried out
by the model.

Complementary representational analysis methods such as Canonical Correlation
Analysis (CCA) \autocite{hotellingRelationsTwoSets1936} and Representational
Similarity Analysis (RSA)
\autocite{kriegeskorteRepresentationalSimilarityAnalysisconnecting2008} are not
directional and sometimes do not even require learnable parameters. Research
using these methods often yields results corroborating those of probing on the encoding of
specific features in the hidden layers of language models
\autocite{chrupalaCorrelatingNeuralSymbolic2019,pasadLayerwiseAnalysisSelfsupervised2021,deheerklootsWhatSelfsupervisedSpeech2025},
but does not address the two main challenges we focus on here (i.e.\ comparison
of different features and controlling of correlation between features).

\subsection{Encoding of speaker identity}
\label{sec:lit-speaker}
Probing experiments have shown that various features can be decoded from the
internal representations of models such as wav2vec2
\autocite{baevskiWav2vec20Framework2020}, including phonetic
\autocite{maProbingAcousticRepresentations2021,delafuenteLayerwiseAnalysisMandarin2024,cormacenglishDomainInformedProbingWav2vec2022}
and prosodic features \autocite{bentumWordStressSelfsupervised2025} as well as
speaker identity \autocite{chiuLargeScaleProbingAnalysis2026}.

Human listeners process speaker identity separately from linguistically relevant
information \autocite{leeAcousticVoiceVariation2019}, suggesting that speech models
may also disentangle representations of these two information types.
\textcite{liuSelfsupervisedPredictiveCoding2023} and
\textcite{gubianAnalyzingRelationshipsPretraining2025} show evidence that while
speaker identity can be decoded from the wav2vec2 model's internal hidden
states, the subspace it occupies is orthogonal to that of phonetic and tonal
information. Our \textsc{Encoding Probe} can provide further insight by directly
comparing the relative contribution of phonetic and speaker features.

\subsection{Encoding of syntactic structure}
\label{sec:lit-syntax}
In the text domain, simple syntactic properties (such as POS labels, tree depth,
constituent labels) have been probed for in neural representations since the
early work of \textcite{adiFinegrainedAnalysisSentence2017} and
\textcite{conneauWhatYouCan2018}. More complex syntactic and semantic structures
have been shown to be recoverable from the hidden state activations of models
such as BERT, relying on a variety of custom methods
\autocite{hewittStructuralProbeFinding2019,chrupalaCorrelatingNeuralSymbolic2019,tenneyBERTRediscoversClassical2019}.
Using a probe formulated to specifically reveal relationships between tokens,
\textcite{hewittStructuralProbeFinding2019} find that the structural
relationships between two token representations are highly correlated with the
structure of a syntax tree. Further experiments using the same structural probe
with curated datasets show that while it is clear that there is some
\emph{syntactic} knowledge encoded by the BERT model, it is difficult to
disentangle syntax from other confounding signals such as semantics or the
closeness of two words in a sentence
\autocite{maudslaySyntacticProbesProbe2021,simon2025probing}.

In the speech domain, pre-trained speech models such as wav2vec2
\autocite{baevskiWav2vec20Framework2020} and HuBERT
\autocite{hsuHuBERTSelfSupervisedSpeech2021} have been shown to encode syntactic
information and word identity
\autocite{pasadLayerwiseAnalysisSelfsupervised2021,shenWaveSyntaxProbing2023,pasadWhatSelfSupervisedSpeech2024,choiSelfSupervisedSpeechRepresentations2024}.
However, \textcite{shenWaveSyntaxProbing2023} note that the syntactic and
lexical features are highly correlated, and the syntactic decodability shown
could possibly be attributed to the underlying lexical representations in the
model embeddings. They come to this conclusion by comparing their syntax probe
when applied to the target model representation versus when applied to a
bag-of-words reference representation. The \textsc{Encoding Probe} offers a more
general way of accounting for such confounds.

\section{Methodology}
\label{sec:method}
While the \textsc{Encoding Probe} can be applied to any target model, in this
paper we focus on transformer language models of text and speech as our main
case studies. We compile a variety of interpretable feature sets that can be
obtained from the speech audio waveform or text input. We then use subsets of
these features to train multivariate ridge regression probes to reconstruct the
hidden state activations of the target models.

\subsection{Models}
We focus on a number of classic Transformer models that have been extensively
used in previous probing studies. We use the model versions listed in
\Cref{tab:models-tested}, which cover different
modalities (text, speech) and training objectives (pre-trained only,
ASR tuning, speaker identification tuning).
\begin{table}[tbp]
  \centering
  \begin{tabular}{ll}
    \toprule
    Model           & Training Objective     \\
    \midrule
    wav2vec2 (base) & Audio self-supervision \\
    wav2vec2 (ASR)  & Speech Recognition     \\
    wav2vec2 (SID)  & Speaker identification \\
    BERT (base)     & Text self-supervision  \\
    \bottomrule
  \end{tabular}
  \caption{Target models tested in this paper.}
  \label{tab:models-tested}
\end{table}
Below we describe the basic
architectures and variants.

\paragraph{Wav2vec2}\autocite{baevskiWav2vec20Framework2020} is a popular speech
model with 12 transformer layers and 768 hidden dimensions, pre-trained on 960
hours of the LibriSpeech dataset \autocite{panayotovLibrispeechASRCorpus2015} in
a self-supervised fashion. We also test the variant fine-tuned for automatic
speech recognition.
In addition to publicly available models, we also fine-tune a wav2vec2-base
model for speaker identification using the 100-hour train split of LibriSpeech
with the \verb|transformers| library
\autocite{wolfTransformersStateoftheArtNatural2020}.
\paragraph{BERT}\autocite{devlin-etal-2019-bert} is a classic text transformer
model. We use \verb|bert-base-uncased|, pre-trained on 3.3B words from
books and web data.

\subsection{Feature sets}
\label{sec:feature-set}

The following interpretable feature sets are used in our experiments, each
representing aspects of human language in the audio and textual modalities.

\paragraph{Acoustics.} Features directly extractable from the waveform signal
\autocite{eybenOpensmileMunichVersatile2010}, including pitch, formants, and
MFCCs\footnote{The complete list of low-level descriptors from the eGeMAPSv02
  feature set \autocite{eybenGenevaMinimalisticAcoustic2016} can be found in
  \Cref{sec:appendix-acoustic-descriptors}}.

\paragraph{Phonetics.} Phonetic posteriorgram (PPG) vectors\footnote{The choice
  of feature representation can affect both probe directions. See
  \Cref{sec:feature-representation} for an analysis of feature representation
  choice using phonetic features as an example.}
\autocite{hazenQuerybyexampleSpokenTerm2009}, representing the probability of
each phone label at a given point in the waveform. We use the \verb|ppgs|
Python package \autocite{churchwellHighFidelityNeuralPhonetic2024} to
automatically generate the PPGs from the waveform.

\paragraph{Speaker.} Speaker identity is an utterance-level label. Within the
speech signal, it is perceived as the quality of the voice or timbre. We one-hot
encode the speaker label from the LibriSpeech dataset
\autocite{panayotovLibrispeechASRCorpus2015} into a feature vector of speaker
identity.

\paragraph{Syntax.} Syntactic features are the grammatical and structural
properties of individual segments of the input. We use spaCy
\autocite{honnibalSpaCyIndustrialstrengthNatural2020} to automatically extract
both categorical features (part-of-speech, dependency label) and numerical
features (total syntax tree depth, depth of token within syntax tree, sequential
position of the word, total number of words in the utterance). The categorical
features are encoded into one-hot vectors.

\paragraph{Lexicon.} Lexical information represents the form and meaning of the
words in the utterance. We use fastText static word embeddings
\autocite{bojanowskiEnrichingWordVectors2017} as a proxy for it. These
embeddings are non-contextual and thus do not contain sentence-level structural
information.

\paragraph{Full.} The combination of all features listed above. The ablations
involve removing features from the union of all the feature sets, e.g.\
$\textit{Full}\setminus\textit{Syntax}$.\footnote{Results of correlation
  checks between feature sets can be found in
  \Cref{sec:corr-speaker,sec:corr-syntax}.}

\subsection{Token and frame embeddings}
In classic text probing studies, token or word embeddings are the base units of
information encoding for target models
\autocite{hewittStructuralProbeFinding2019}.
When probing speech models,
we work directly at
the level of frames, which are the basic units extracted from the audio and
which the transformer layers of speech models use as tokens. We randomly sample
frames from a given utterance and look up the corresponding features for that
given timestamp. The features themselves are computed on frames, words or whole
utterances, as appropriate, but the model representations always correspond to
single frames. This obviates the need to pool representations, and minimizes the
imposition of text-based concepts on the analysis of models of spoken language.

Word-level annotations (syntactic features and fastText embeddings) are attached
to frames and subword tokens using forced-alignment word timestamps generated
with WebMAUS \autocite{kislerMultilingualProcessingSpeech2017}. For each sampled
frame (in speech models) or subword token (in BERT), we look up the
corresponding word segment and assign that word's annotation. We thus probe the
frame-level representations of wav2vec2 and the subword tokens of BERT directly,
without pooling multiple frames into word-level representations.

\subsection{The \textsc{Encoding Probe}}

We fit a Ridge regressor with \verb|scikit-learn|
\autocite{pedregosaScikitlearnMachineLearning2011} that takes the union of all
feature sets \textit{Full} to predict the model representation of the
corresponding frame. Essentially, the probe learns to explain the variance
within the model representations, and is evaluated on the test split of the
experimental data. As our reconstruction error metric, we report the
\textbf{unexplained variance} (UV) of the probe, defined as the ratio of
residual sum of squares to the total sum of squares of the dependent variable:
\begin{equation}
  \text{UV} = \frac{SS_{\text{res}}}{SS_{\text{tot}}} = 1 - R^2.
\end{equation}
After a feature is ablated, a larger upward gap from the \textit{Full} baseline
indicates a greater relative contribution for that feature, irrespective of
absolute UV values. The regularization strength $\alpha$ was tuned via
cross-validation over the grid $\{10^n \mid n \in \mathbb{Z}, -3 \le n \le 5\}$.
We refit the ridge regressor with the same hyperparameter search for every
ablation.

We use the 100-hour train split of the LibriSpeech dataset
\autocite{panayotovLibrispeechASRCorpus2015} and employ a randomized 80:20
train-test split stratified by speaker ID. Stratification is applied only to
speaker ID since it is the only utterance-level label we include in the probe's
input. We select a maximum of 10 subword tokens from each utterance in text
models and a maximum of 15 frames from speech models. Silent frames are filtered
out. The preprocessing pipeline results in ${\sim}230$k input-output pairs for
speech models and ${\sim}180$k for text models. To quantify the impact of
sampling frames, we ran the experiments for \verb|wav2vec2-base| and
\verb|bert-base-uncased| with 10 additional random seeds $\{n\times 100 \mid n
  \in \mathbb{Z}, 1 \le n \le 10\}$ and found that the 95\% confidence interval
(CI) width of the mean differences of UV scores across all random seeds was
$<0.002$.

\paragraph{A built-in comparison point.} We compare each probe to the analogous
one with the complete set of features, \textit{Full}, which by design yields the
lowest Unexplained Variance (UV).

\paragraph{Parallel decoding probe analyses.} We also implement parallel
analyses with the \textsc{Decoding Probe} alongside the \textsc{Encoding Probe}
results. Our decoding analyses use \verb|scikit-learn| Ridge regressors and
classifiers that decode features of interest using the frame-level
representations of the transformer models. We utilize the same features across
both probe types, except that class labels are used instead of one-hot encoded
vectors for the classification tasks. We conduct a hyperparameter search using
the same parameter grid as above.

\subsection{Design of case studies}
\label{sec:experimental-design}
We present two case studies to demonstrate the application of the
\textsc{Encoding Probe}: the encoding of speaker identity in models of speech,
and the encoding of syntactic features in models of text and speech. Both of
these topics have been extensively investigated in prior research as described
in \Cref{sec:literature}, where the \textsc{Decoding Probe} has been used to
decode relevant features from various transformer models' inner representations.
As discussed in \Cref{sec:decoding-probe-limitations}, in both cases there is
another, potentially highly correlated source of information (acoustics in the
case of speaker identity and lexical features in the case of syntactic features)
which might affect the previously reported findings. We design two sets of
experiments for each case study, using a set of targeted transformer models to
demonstrate how the \textsc{Encoding Probe} offers a complementary perspective
that mitigates limitations of the \textsc{Decoding Probe}.

\section{Experiment 1: Speaker Identity}
\label{sec:speaker-identity}

Here we investigate the relative contribution of speaker identity, phonetic, and
acoustic features. We also look at the impact of
the tuning objectives of the models on the encoding of these features in their internal representations.

\subsection{Decoding speaker identity and phone labels}
\begin{figure}[tbp]
  \centering
  \includegraphics[width=\linewidth]{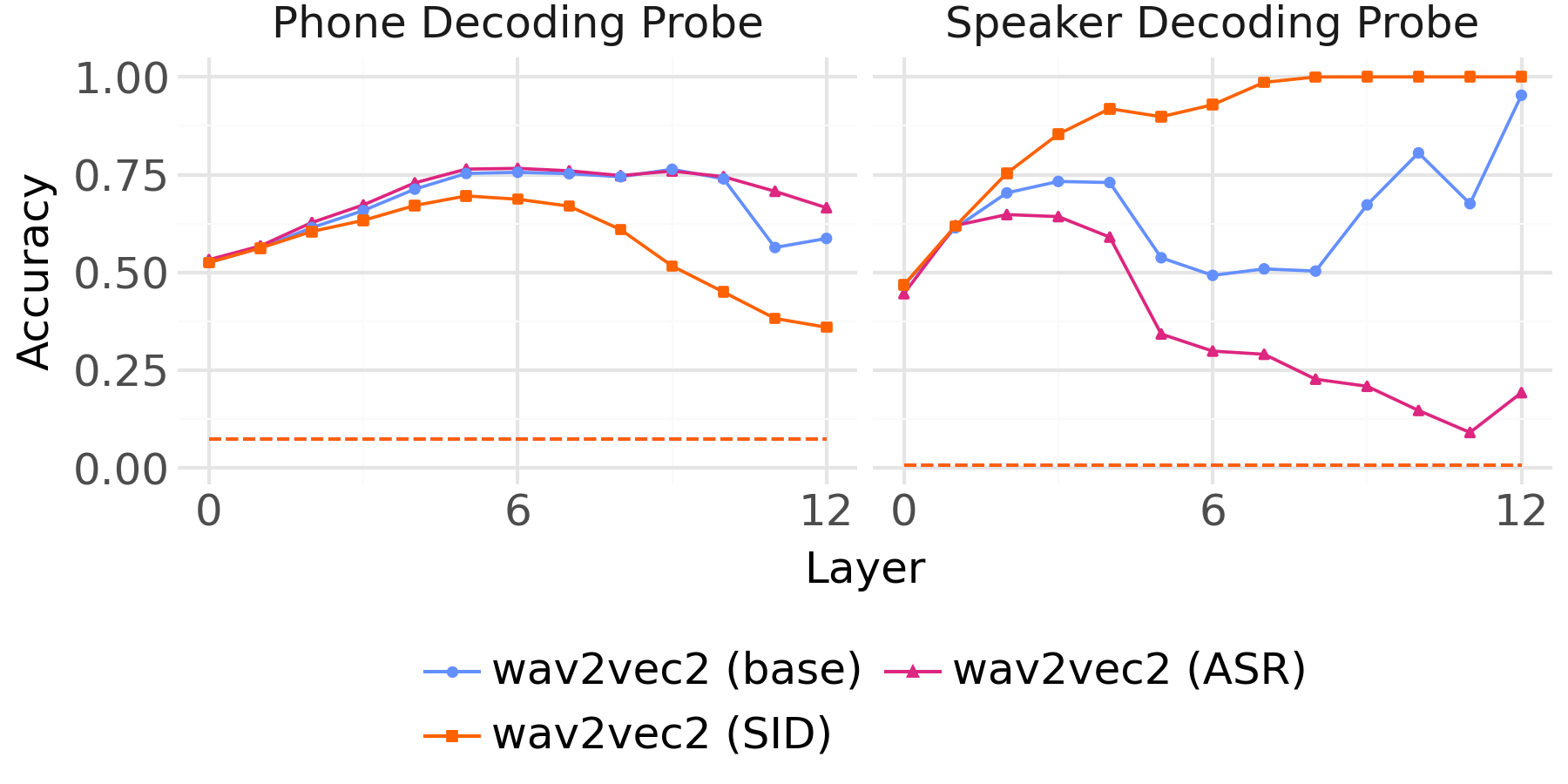}
  \caption{Accuracy of the \textsc{Decoding Probe} on predicting speaker
    identity and phone label from frames of model internal hidden states.
    The dashed line represents the majority baseline of the classification
    task.}
  \label{fig:wav2vec2-dec-speaker-phonetics}
\end{figure}

Before applying the \textsc{Encoding Probe} for this purpose, we first use
the \textsc{Decoding Probe} to establish a point of comparison for speaker identity
and phone labels. We fit linear models to decode speaker identity and phone
label from the hidden-state representations of the speech models. As shown in
the right panel of \Cref{fig:wav2vec2-dec-speaker-phonetics}, we find the
performance of the speaker identity probe increases towards later layers in the
base and SID models but decreases in the ASR model (with a dip in performance in
the middle layers for base). The left panel of
\Cref{fig:wav2vec2-dec-speaker-phonetics} shows the phone label probe
performance increases from the early to middle layers but drops in final layers
for the base and SID models and remains high in the ASR model. However, the
performance of all probes consistently exceeds the majority baseline for all
layers in all models.

These results show that the decodability of speaker identity and phone labels
varies depending on the training/tuning objectives of the model. The tuning
objective of the ASR model pushes the model to normalize across different
speakers while maintaining its ability to identify the correct phone label for
the CTC decoding task, whereas the SID model's objective encourages it to
maximize its ability to differentiate speakers, resulting in a larger drop in
the decodability of phone labels.

These patterns in the current findings confirm previous research. However, the
results only show that there is enough information in the model's internal
hidden states for the probe to decode these features. Due to differences in the
baselines and the levels of difficulty of the tasks, the results of these two
probes are not directly comparable against each other, and the relative
contribution of each feature to model representations cannot be quantified.

\subsection{Encoding to reconstruct representations}

Now we apply the \textsc{Encoding Probe} to reconstruct representations
of the same speech models (base, ASR-tuned, and SID-tuned variants of
wav2vec2)\footnote{The experimental results for an extended set of models can be
  found in \Cref{sec:all-model-results}.}. In addition to the \textit{Full} probe, we also
construct ablated feature sets with each of the targeted feature sets
(\textit{Acoustics, Phonetics, Speaker}) and joint ablations of two sets.

\begin{figure}[tbp]
  \centering
  \begin{subfigure}{\linewidth}
    \centering
    \includegraphics[width=\linewidth]{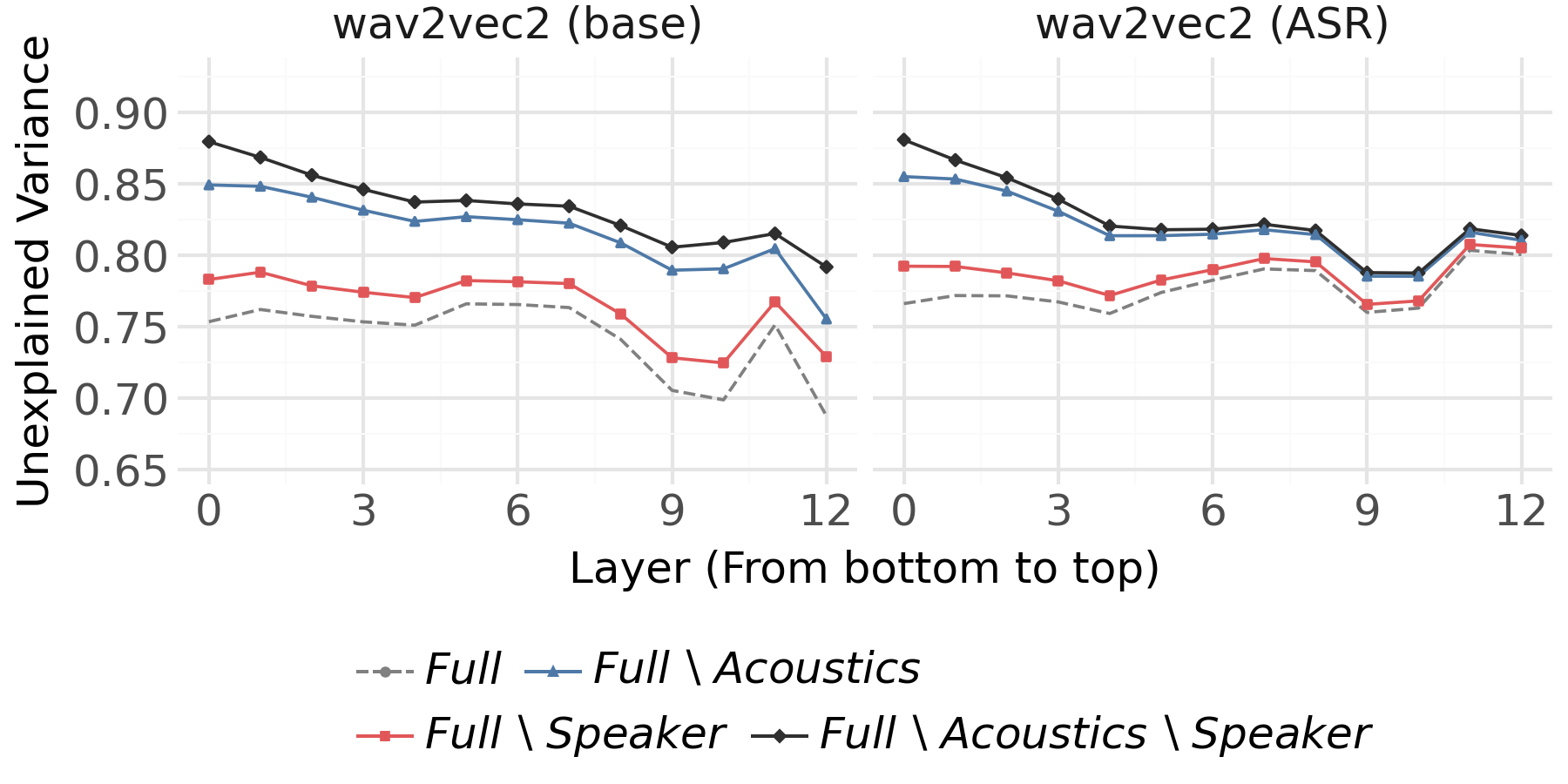}
    \caption{Acoustics and speaker identity}
    \label{fig:acoustic-speaker-2}
  \end{subfigure}
  \begin{subfigure}{\linewidth}
    \centering
    \includegraphics[width=\linewidth]{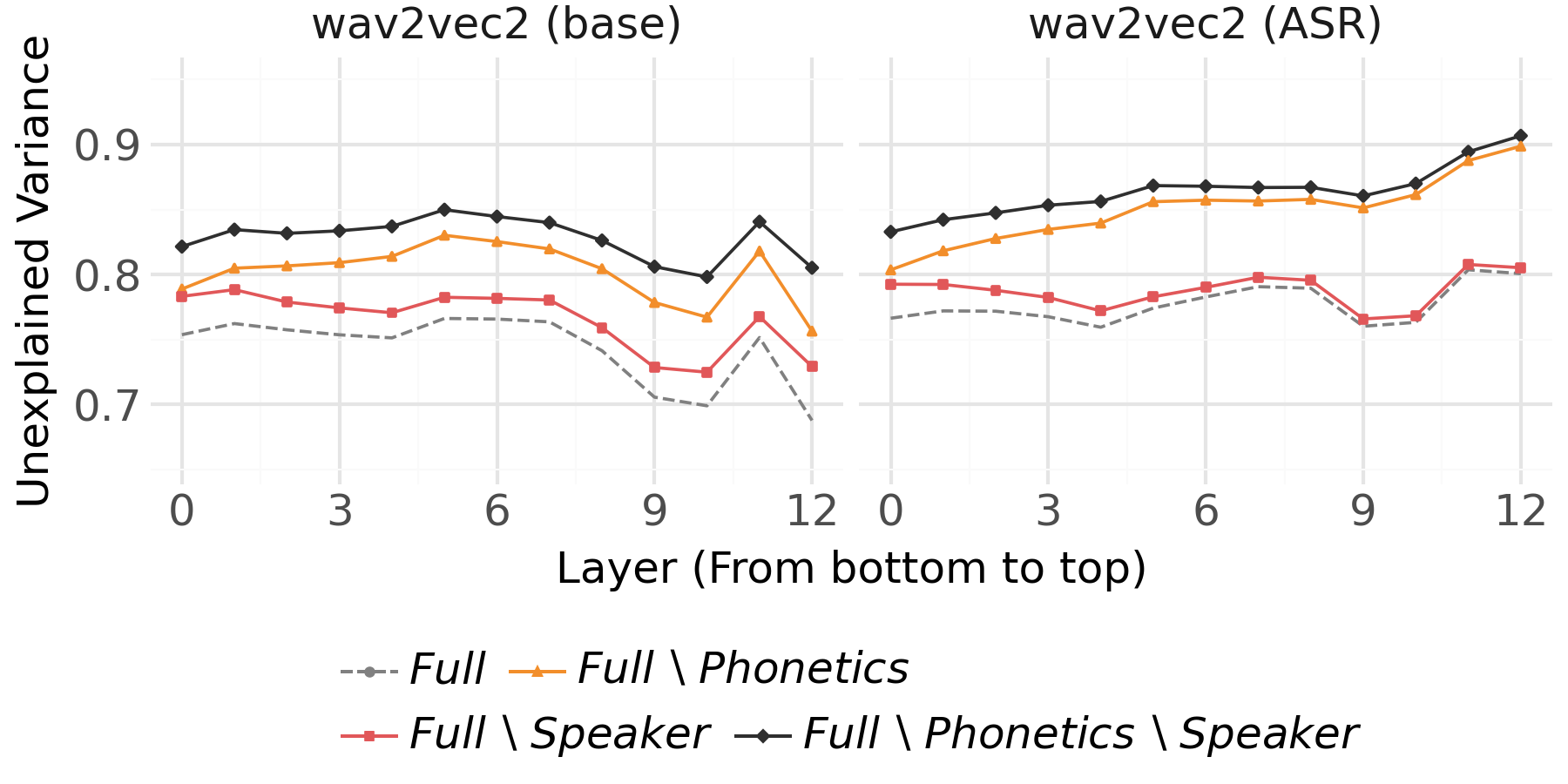}
    \caption{Phonetics and speaker identity}
    \label{fig:phonetic-speaker-2}
  \end{subfigure}
  \caption{Unexplained Variance (UV) for the base and ASR models (larger gap
    from \textit{Full}/dashed = greater contribution).}
  \label{fig:wav2vec-enc-base-asr}
\end{figure}

\begin{figure}[tbp]
  \centering
  \begin{subfigure}{\linewidth}
    \centering
    \includegraphics[width=\linewidth]{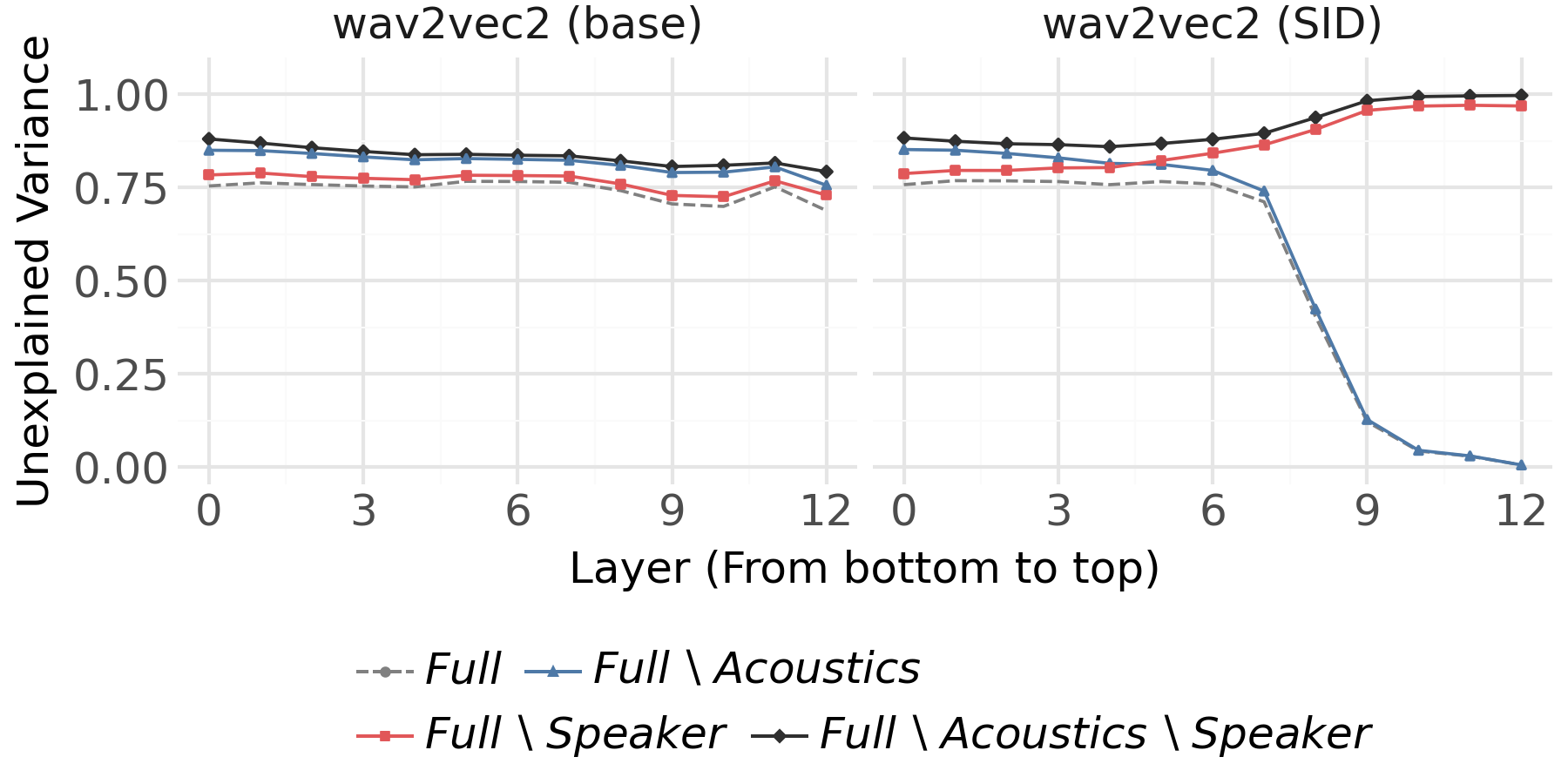}
    \caption{Acoustics and speaker identity}
    \label{fig:acoustic-speaker-1}
  \end{subfigure}
  \begin{subfigure}{\linewidth}
    \centering
    \includegraphics[width=\linewidth]{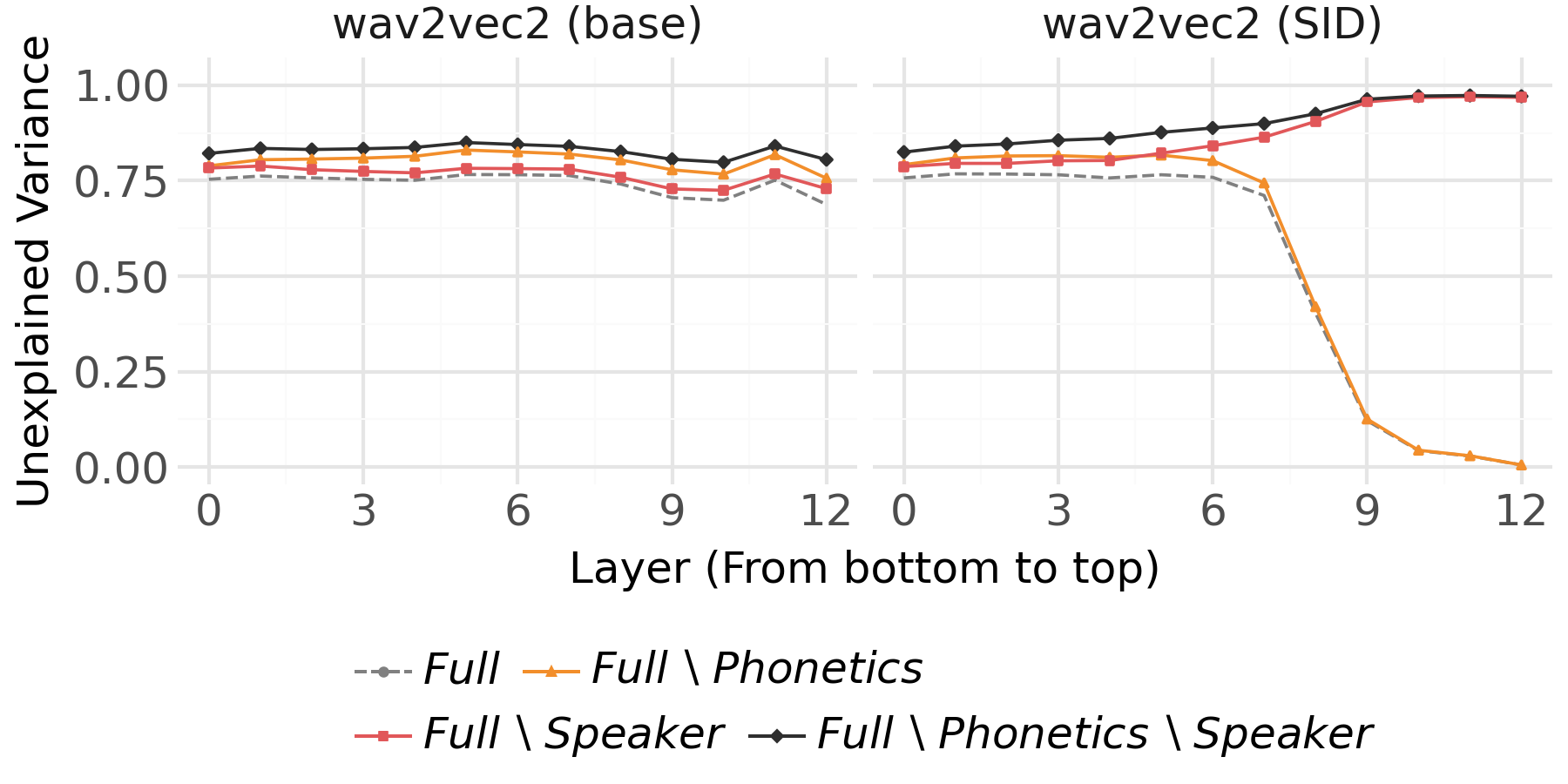}
    \caption{Phonetics and speaker identity}
    \label{fig:phonetic-speaker-1}
  \end{subfigure}
  \caption{Unexplained Variance (UV) for the base and SID models (larger gap
    from \textit{Full}/dashed = greater contribution).}
  \label{fig:wav2vec-enc-base-sid}
\end{figure}

\Cref{fig:wav2vec-enc-base-asr,fig:wav2vec-enc-base-sid} show the
\textsc{Encoding Probe} results for the speaker identity experiments (note that
the two sets of graphs use different scales on their y-axes). A larger gap
between the grey dashed line (i.e.\ the \textit{Full} probe) and a colored solid
line (i.e.\ an ablated probe) means that the ablated feature has a larger
relative contribution to the reconstruction of the representation.


We first focus on \Cref{fig:wav2vec-enc-base-asr}, which shows the impact of
speaker identity, phonetic, and acoustic features in the base and ASR-tuned
models. The top row shows that the UV of the
$\textit{Full}\setminus\textit{Acoustics}$ probe is substantially higher than
that of the $\textit{Full}\setminus\textit{Speaker}$ probe, for both models and
all layers (blue line is always above red line). The bottom row shows that
$\textit{Full}\setminus\textit{Phonetics}$ also scores higher than
$\textit{Full}\setminus\textit{Speaker}$. These combined results indicate that
both acoustic and phonetic features have a larger relative feature contribution
than speaker identity in the base and ASR-tuned models. Comparing the results
between the left and right columns, we see a smaller gap between
$\textit{Full}\setminus\textit{Speaker}$ and \textit{Full}. This indicates that
the relative contribution of speaker identity is smaller in the ASR-tuned model
than in the base model. Additionally, in the upper layers, we can see a larger
effect for $\textit{Full}\setminus\textit{Phonetics}$ for the ASR-tuned model
compared to the base model, whereas $\textit{Full}\setminus\textit{Acoustics}$
shows the opposite relationship, indicating that phonetic features are encoded
more prominently in the internal hidden states of the ASR-tuned model than in
those of the base model.

For the comparison between base and SID-tuned models shown in
\Cref{fig:wav2vec-enc-base-sid}, the situation is similar in the bottom layers,
but changes drastically in the upper layers. First, for layers 9-12 of
the SID-tuned model, \textit{Full} achieves a very low UV, i.e.\ it accounts for
most of the variance in the target model representations. Furthermore, from
around layer~7 of the SID model onward, both
$\textit{Full}\setminus\textit{Acoustics}$ and
$\textit{Full}\setminus\textit{Phonetics}$ barely show any increase above
\textit{Full}. In the same layers the UV of the
$\textit{Full}\setminus\textit{Speaker}$ probe, on the other hand, is close to
one. This profile of results indicates that the top layers of the SID-tuned
model are almost exclusively dedicated to encoding speaker identity to the
exclusion of any other information.

\subsection{Decoding versus encoding results}
To highlight the different insights obtainable via the \textsc{Decoding Probe}
and \textsc{Encoding Probe}, we focus on the especially striking results for the
tuned models. We can decode phonetics from the representations in the top layers
of the SID model with non-trivial accuracy, while at the same time phonetic
features barely contribute to the reconstruction of these representations.
Similarly, we can decode speaker labels from the top layers of the ASR-tuned
model with above-baseline accuracy, but the contribution of speaker identity to
the reconstruction of these same layers is very minor. In other words,
\textsc{Decoding Probe} detects phonetic structure sufficient for
classification, but the \textsc{Encoding Probe} reveals that this structure
accounts for a negligible fraction of the total representational variance in
these layers. The fact that decodability and relative contribution can diverge
substantially underscores the value of the \textsc{Encoding Probe}.

\section{Experiment 2: Syntactic Information}
\label{sec:syntax}

In this section we quantify the contributions of syntactic versus lexical
information to the reconstruction of target models' representations, while
controlling for correlations between these features, which have been highlighted
in previous work. We examine both text and speech transformer models.

\subsection{Decoding syntactic and lexical features}
\begin{figure}[tbp]
  \centering
  \includegraphics[width=\linewidth]{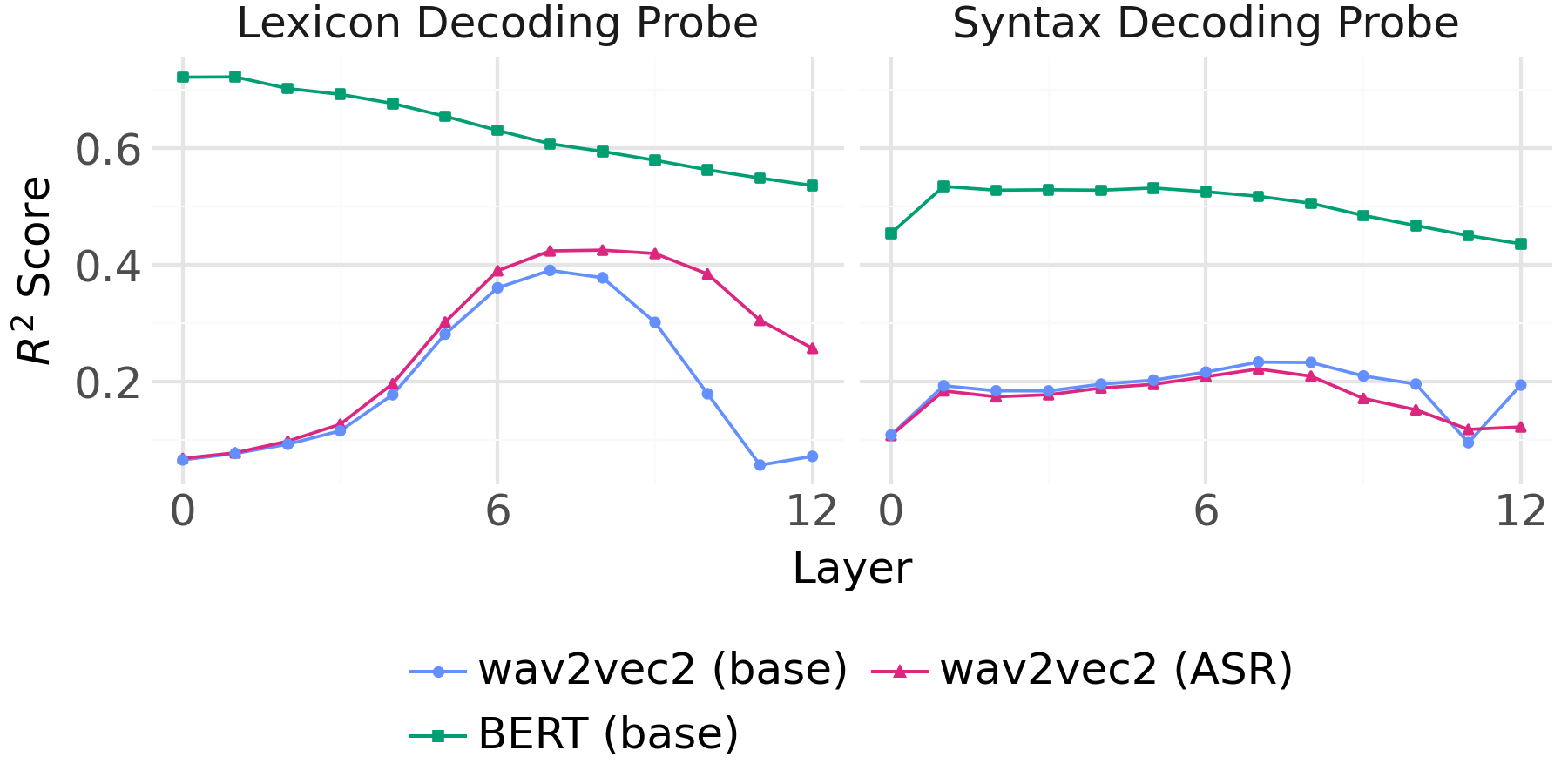}
  \caption{$R^2$ of the \textsc{Decoding Probe} on predicting syntactic and
    lexical feature vectors from the frames of target model representations.}
  \label{fig:wav2vec2-dec-syntax-lexicon}
\end{figure}

We first apply a \textsc{Decoding Probe} to establish the decodability of
syntactic and lexical features separately.
\Cref{fig:wav2vec2-dec-syntax-lexicon} shows that syntactic and lexical features
are much more decodable in BERT (base) than in the wav2vec2 models. However,
while the highest $R^2$ score of the probes for both the base and ASR-tuned
models appears around layer~7, the pattern from layers 9-12 is slightly different.
The lexical probe for the ASR-tuned model has a smaller decrease in $R^2$ score
compared to the probe for the base model, and the syntax probes show smaller
differences between the two models.

\subsection{Reconstructing representations}
\begin{figure}[tbp]
  \centering
  \begin{subfigure}{\linewidth}
    \centering
    \includegraphics[width=\linewidth]{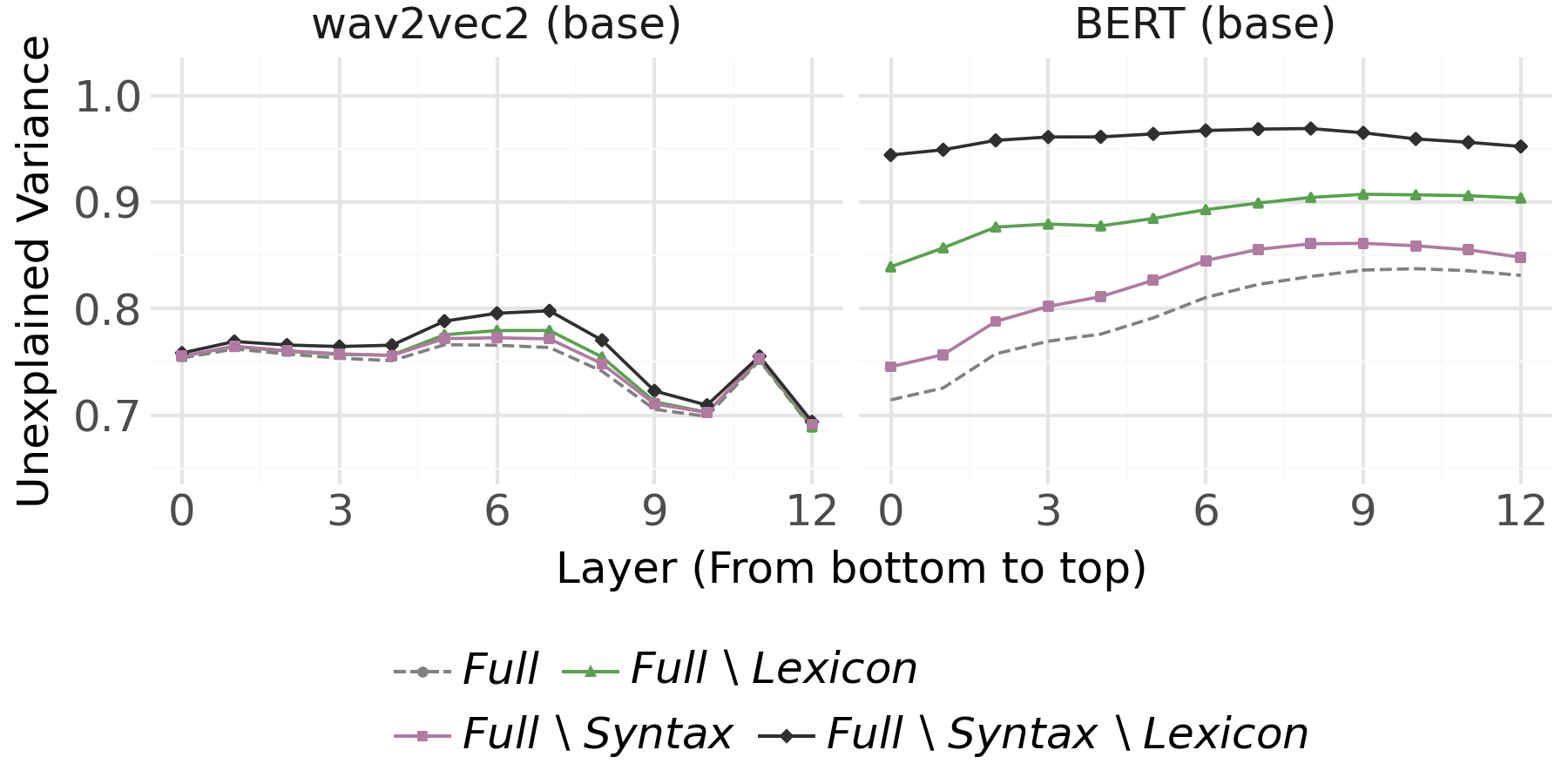}
    \caption{Speech versus text}
    \label{fig:enc-syntax-aud-text}
  \end{subfigure}
  \begin{subfigure}{\linewidth}
    \centering
    \includegraphics[width=\linewidth]{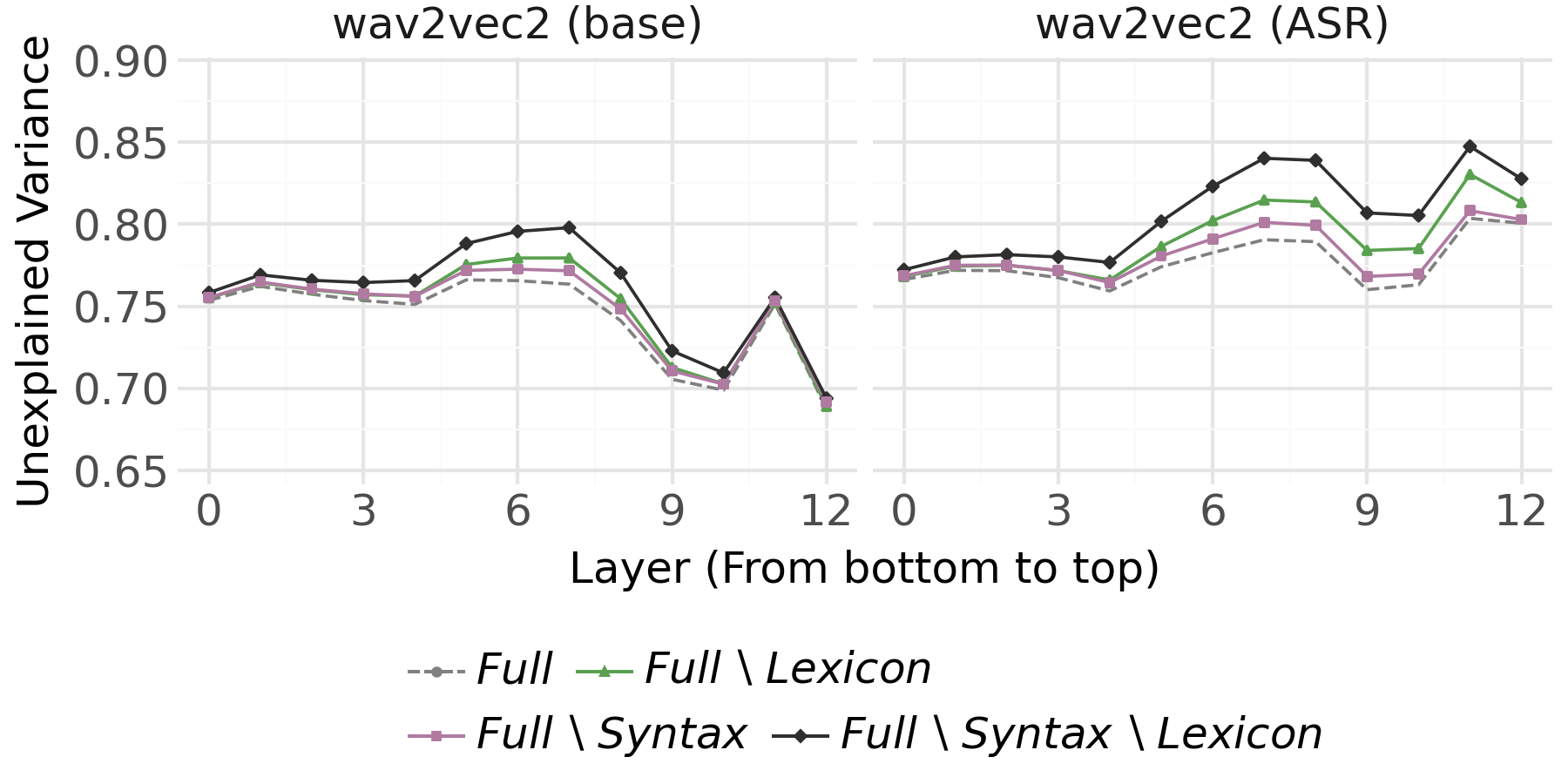}
    \caption{Base versus ASR-tuned}
    \label{fig:enc-syntax-ft}
  \end{subfigure}
  \caption{Unexplained Variance (UV) $(1-R^2)$ of the \textsc{Encoding Probe}
    for the text base, speech base, and ASR models (larger gap from
    \textit{Full}/dashed = greater contribution).}
  \label{fig:enc-syntax}
\end{figure}

\Cref{fig:enc-syntax} shows the results of applying the \textsc{Encoding Probe}
to reconstruct the representations of one text model, BERT (base), and two
speech models (base and ASR variants of wav2vec2)\footnote{The experimental
  results for an extended set of models can be found in
  \Cref{fig:all-model-syntax} in the Appendix.}. We re-use the \textit{Full} probe
and report additional ablated feature sets with each of the targeted features
(\textit{Syntax, Lexicon}) as well as joint ablation of two sets. We discuss the
main observations below.

First, the increase in UV due to ablating lexical features is higher than the
increase due to ablating syntactic features: the line corresponding to
$\textit{Full}\setminus\textit{Lexicon}$ is consistently above
$\textit{Full}\setminus\textit{Syntax}$, for all three target models.

Second, the ablated probes always have higher UV scores than the \textit{Full}
probe, which indicates that lexical and syntactic features each make an
independent contribution in the probe's reconstruction of model activation
patterns. The hypothesis that syntactic decodability is entirely attributable to
lexical correlations is not supported: the ablation of syntactic features
produces an increase in UV even when lexical features are present in the probe
input.

Regarding the impact of the target model modality, the effect of the ablations
is consistently larger in BERT (base) than in the speech models, meaning that lexical
and syntactic features make a more prominent contribution to the reconstruction of
the representations of BERT (base) compared to those of the speech models.

Finally, fine-tuning on ASR leads to a target model whose representations are
somewhat harder to reconstruct in absolute terms, but where lexical and
syntactic features contribute more to the reconstruction compared with what they
contribute to the base model.

\subsection{Decoding versus encoding results}
The \textsc{Decoding Probe} shows that lexical and syntactic features are
decodable with moderate success, but only the \textsc{Encoding Probe} allows
quantitative comparison of their relative contributions while controlling for
correlations. This case illustrates that the \textsc{Decoding Probe} shows whether
information is present, and the \textsc{Encoding Probe} reveals how much of the
representational variance it explains.

\section{Discussion and Conclusion}
\label{sec:conclusion}

We present the \textsc{Encoding Probe}, which reconstructs representations in
the layers of a target model based on multiple interpretable features. Our probe
enables direct comparison of the relative contributions of different features
through a feature ablation analysis under a shared regression objective, while
controlling for correlated features.

We design two case studies to showcase the \textsc{Encoding Probe}. In the
first, speaker identity is decodable across all wav2vec2 variants, confirming
earlier findings using decoding probes
(\textcite{gubianAnalyzingRelationshipsPretraining2025}), yet its contribution
to reconstruction is modest in base and ASR models and substantial only with
speaker-ID tuning. In the second, lexical and syntactic features make
independent contributions to reconstruction in both text and speech models, with
lexical features accounting for a higher fraction of the variance, confirming
that these models encode syntactic information beyond what lexical features
alone can explain.

Such observations are only possible because the multiple regression analysis
framework underlying the \textsc{Encoding Probe} enables us to control for
correlations between features by design. This framework likewise leads to a
direct quantification of relative feature contributions.

\section*{Limitations}
The \textsc{Encoding Probe} addresses some of the issues with probing
highlighted by \textcite{belinkovProbingClassifiersPromises2022}, but new
limitations arise and some are not addressed. The probe remains observational
rather than interventional: it reveals statistical associations between
interpretable features and model representations, but it does not establish
causal mechanisms. While the \textsc{Encoding Probe} framework in itself
supports the use of any regression model, our implementation here relies on
ridge regression, which may fail to capture nonlinear relationships in the
transformer representations. While nonlinear probes might achieve lower absolute
reconstruction error, linear probes offer better interpretability and are
standard in the probing literature. The encoding results are also conditional on the
feature sets we chose to operationalize the input space (i.e., acoustics,
phonetics, syntax, lexicon, and speaker identity).  Factors not included in our
analysis can still contribute to the probe's reconstruction performance. The
measured contribution of each feature block also depends on its dimensionality
and richness, so the cross-feature rankings should be interpreted cautiously
(\Cref{sec:pca-matched-control}). While
the frame-level setup allows for a faithful analysis of the representations of
speech models, it potentially under-represents long-range structural
information. Our speech experiments use read audiobook data; results may differ
for spontaneous or noisy speech. Finally, observed feature contributions and
interaction patterns may vary with dataset composition and alignment quality.
For these reasons, the reported ablation effects should be read as a case study,
rather than as an exhaustive characterization of internal representations of the
target models tested.

\section*{Acknowledgments}

This publication is part of the project \textit{InDeep: Interpreting Deep
  Learning Models for Text and Sound} (with project number NWA.1292.19.399) of
the National Research Agenda (NWA-ORC) program. Funding by the Dutch Research
Council (NWO) is gratefully acknowledged. We also thank SURF (www.surf.nl) for
the support in using the National Supercomputer Snellius.

\appendix
\section{Appendix}
\label{sec:appendix}

\subsection{Acoustic descriptor inventory}
\label{sec:appendix-acoustic-descriptors}
This section outlines the feature names for the complete
eGeMAPSv02\autocite{eybenGenevaMinimalisticAcoustic2016} feature set extracted
by openSMILE \autocite{eybenOpensmileMunichVersatile2010}. We use the low-level
descriptors that extract each descriptor at 20 ms intervals to best align with
the frame sizes of the speech models (25 ms for wav2vec2).
\begin{table}[ht]
  \centering
  \begin{tabular}{ll}
    \toprule
    \# & Acoustic descriptor          \\
    \midrule
    1  & Loudness\_sma3               \\
    2  & alphaRatio\_sma3             \\
    3  & hammarbergIndex\_sma3        \\
    4  & slope0-500\_sma3             \\
    5  & slope500-1500\_sma3          \\
    6  & spectralFlux\_sma3           \\
    7  & mfcc1\_sma3                  \\
    8  & mfcc2\_sma3                  \\
    9  & mfcc3\_sma3                  \\
    10 & mfcc4\_sma3                  \\
    11 & F0semitoneFrom27.5Hz\_sma3nz \\
    12 & jitterLocal\_sma3nz          \\
    13 & shimmerLocaldB\_sma3nz       \\
    14 & HNRdBACF\_sma3nz             \\
    15 & logRelF0-H1-H2\_sma3nz       \\
    16 & logRelF0-H1-A3\_sma3nz       \\
    17 & F1frequency\_sma3nz          \\
    18 & F1bandwidth\_sma3nz          \\
    19 & F1amplitudeLogRelF0\_sma3nz  \\
    20 & F2frequency\_sma3nz          \\
    21 & F2bandwidth\_sma3nz          \\
    22 & F2amplitudeLogRelF0\_sma3nz  \\
    23 & F3frequency\_sma3nz          \\
    24 & F3bandwidth\_sma3nz          \\
    25 & F3amplitudeLogRelF0\_sma3nz  \\
    \bottomrule
  \end{tabular}
  \caption{Full openSMILE eGeMAPSv02 low-level descriptor inventory used for the acoustic feature set in the probe input ensemble.}
  \label{tab:acoustic-descriptor-inventory}
\end{table}

\subsection{Different feature representations}
\label{sec:feature-representation}
\begin{figure}
  \centering
  \includegraphics[width = \linewidth]{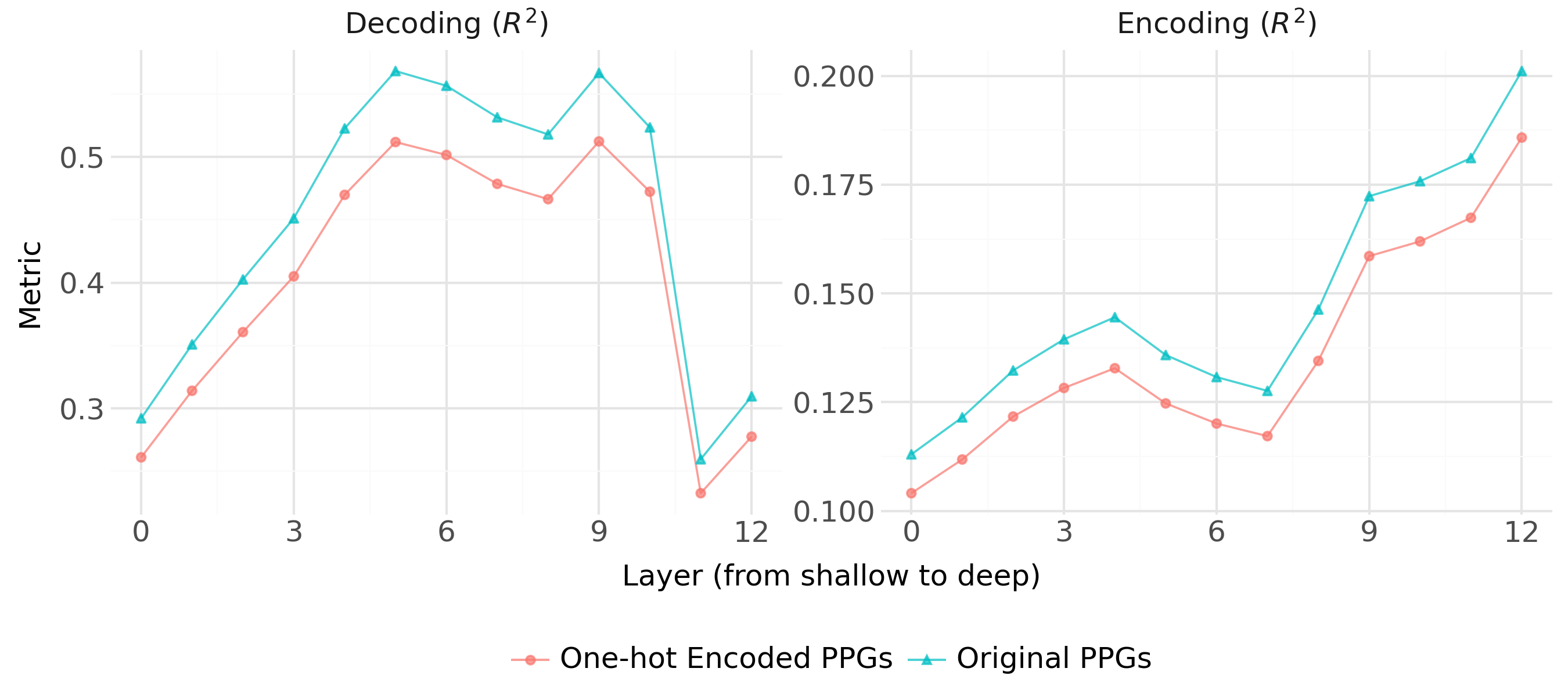}
  \caption{Comparison between \textsc{Encoding Probe} and \textsc{Decoding
      Probe} results using different representations of the same phonetic
    feature.}
  \label{fig:feature_rep}
\end{figure}

To show the importance of feature representation type, we test two different
ways of representing phonetic information in both the \textsc{Encoding Probe}
and \textsc{Decoding Probe} settings: the raw phonetic posteriorgram (PPG) and a
one-hot encoding derived from the PPGs. The results are reported in
\Cref{fig:feature_rep}. The raw PPGs consistently achieve higher $R^2$ than the
one-hot PPGs in the \textsc{Encoding Probe}, which is consistent with the
one-hot encoding discarding additional information from the probability of other
phones present at each timestamp. The \textsc{Decoding Probe} shows the same
ordering of the two representations. Choosing an appropriate feature
representation therefore affects both probe directions, and should be justified
when interpreting probe results.

\subsection{Decoding probe results}
\label{sec:decoding-results}
\begin{table}[!ht]
  \centering
  \caption{Layerwise decoding probe results for wav2vec2-base on the LibriSpeech
    train-clean-100 subset. We report the held-out test accuracy, balanced
    accuracy, and macro-F1 for decoding speaker IDs and phone labels from each
    layer. Bold values mark the layer with the highest balanced accuracy for
    each feature.}
  \label{tab:decoding-results}
  \begin{tabular}{@{}c c c c@{}}
    \toprule
    Layer       & Acc.          & Bal.\ acc.    & Macro-F1      \\
    \midrule
    \multicolumn{4}{@{}l}{\emph{Speaker ID}}                    \\
    0           & 46.7          & 43.7          & 42.4          \\
    1           & 61.4          & 58.5          & 58.1          \\
    2           & 70.3          & 67.5          & 68.2          \\
    3           & 73.3          & 70.4          & 71.5          \\
    4           & 73.0          & 69.8          & 70.8          \\
    5           & 53.8          & 50.1          & 50.4          \\
    6           & 49.3          & 45.7          & 45.7          \\
    7           & 50.9          & 47.3          & 47.3          \\
    8           & 50.3          & 46.7          & 46.7          \\
    9           & 67.3          & 63.6          & 63.6          \\
    10          & 80.6          & 77.8          & 78.0          \\
    11          & 67.6          & 64.3          & 67.3          \\
    \textbf{12} & \textbf{95.3} & \textbf{94.5} & \textbf{94.9} \\
    \midrule
    \multicolumn{4}{@{}l}{\emph{Phone label}}                   \\
    0           & 52.5          & 39.9          & 39.5          \\
    1           & 56.3          & 44.0          & 43.9          \\
    2           & 61.5          & 50.1          & 50.6          \\
    3           & 65.8          & 55.6          & 56.6          \\
    4           & 71.3          & 63.1          & 64.6          \\
    5           & 75.3          & 69.7          & 71.5          \\
    6           & 75.6          & 69.8          & 71.7          \\
    7           & 75.3          & 69.0          & 71.1          \\
    8           & 74.5          & 66.7          & 68.7          \\
    \textbf{9}  & \textbf{76.4} & \textbf{70.3} & \textbf{72.2} \\
    10          & 73.9          & 65.5          & 67.4          \\
    11          & 56.4          & 43.3          & 45.7          \\
    12          & 58.7          & 46.3          & 47.2          \\
    \bottomrule
  \end{tabular}
\end{table}
\Cref{tab:decoding-results} reports the held-out decoding probe results used in
the motivating example of \Cref{sec:decoding-probe-limitations}. For each
feature we train a linear decoding probe on every hidden-state layer of
wav2vec2-base and report the test accuracy, balanced accuracy, and macro-F1.
Bold values mark the layer with the highest balanced accuracy for each feature.

\subsection{Direct correlations between features}
\subsubsection{Speaker ID, phonetic, and acoustic features}
\label{sec:corr-speaker}
We directly estimate the linear correlations between speaker identity and
phonetic or acoustic features. We fit a ridge classifier using phonetic,
acoustic and the combined features to predict the speaker identity. The results
in \Cref{tab:input-overlap-spk} show that speaker identity is only weakly
decodable from the explicit acoustic and phonetic features with a linear
classifier. While it is not surprising that there is no correlation between
phonetics and speaker identity, we also could not decode speaker label directly
from acoustic features. Our acoustic features consist of highly localized
acoustic descriptors based on a 20 ms waveform segment. The results show that
there are almost no speaker characteristics that can be extracted from the
\textit{local} acoustic features.

\begin{table}[ht]
  \centering
  \begin{tabular}{ll}
    \toprule
                          & Accuracy \\
    \midrule
    Acoustics             & 0.061    \\
    Phonetics             & 0.008    \\
    Phonetics + Acoustics & 0.071    \\
    \bottomrule
  \end{tabular}
  \caption{Accuracy of a linear classifier using acoustics, phonetics, and combined acoustics + phonetics to predict speaker identity. The majority baseline is around 0.004.}
  \label{tab:input-overlap-spk}
\end{table}

\subsubsection{Syntactic and lexical features}
\label{sec:corr-syntax}
We also directly estimate the linear correlations between syntactic features and
word embeddings. We fit ridge linear regressors/classifiers to predict
individual syntactic features from the static fastText word embeddings. Results
reported in \Cref{tab:input-overlap-lex-syn} indicate that syntactic category
labels (e.g. part-of-speech) are more decodable from the word embeddings than
the other syntactic features. Overall, these results indicate that static word
embeddings are not directly correlated with most of the syntactic features we
include in this study.

\begin{table}[ht]
  \centering
  \begin{tabular}{lll}
    \toprule
                     & Metric   & Score \\
    \midrule
    POS              & Accuracy & 0.628 \\
    Dependency label & Accuracy & 0.417 \\
    Tree depth       & $R^2$    & 0.030 \\
    Word position    & $R^2$    & 0.012 \\
    Total tree depth & $R^2$    & 0.012 \\
    Total word count & $R^2$    & 0.009 \\
    \bottomrule
  \end{tabular}
  \caption{Linear probe results on decoding syntactic features from fastText word embeddings.}
  \label{tab:input-overlap-lex-syn}
\end{table}

\subsection{Ablation and permutation controls}
\label{sec:ablation-controls}
Our experimental paradigm primarily used ablation to remove features from the
input ensemble of the \textsc{Encoding Probe}. However, since ablation changes
the input dimension, it is possible that some of the gaps in the unexplained
variance metric reported in \cref{sec:speaker-identity,sec:syntax} can instead
be attributed to the smaller input dimension. To control for this, we ran a
parallel experiment using permutation instead of ablation. By permuting
(shuffling) the target feature block along the sample axis, we preserve the
input dimension but remove any potential alignment between the content inside
the feature block and the real data representations.

\begin{figure}
  \centering
  \includegraphics[width = \linewidth]{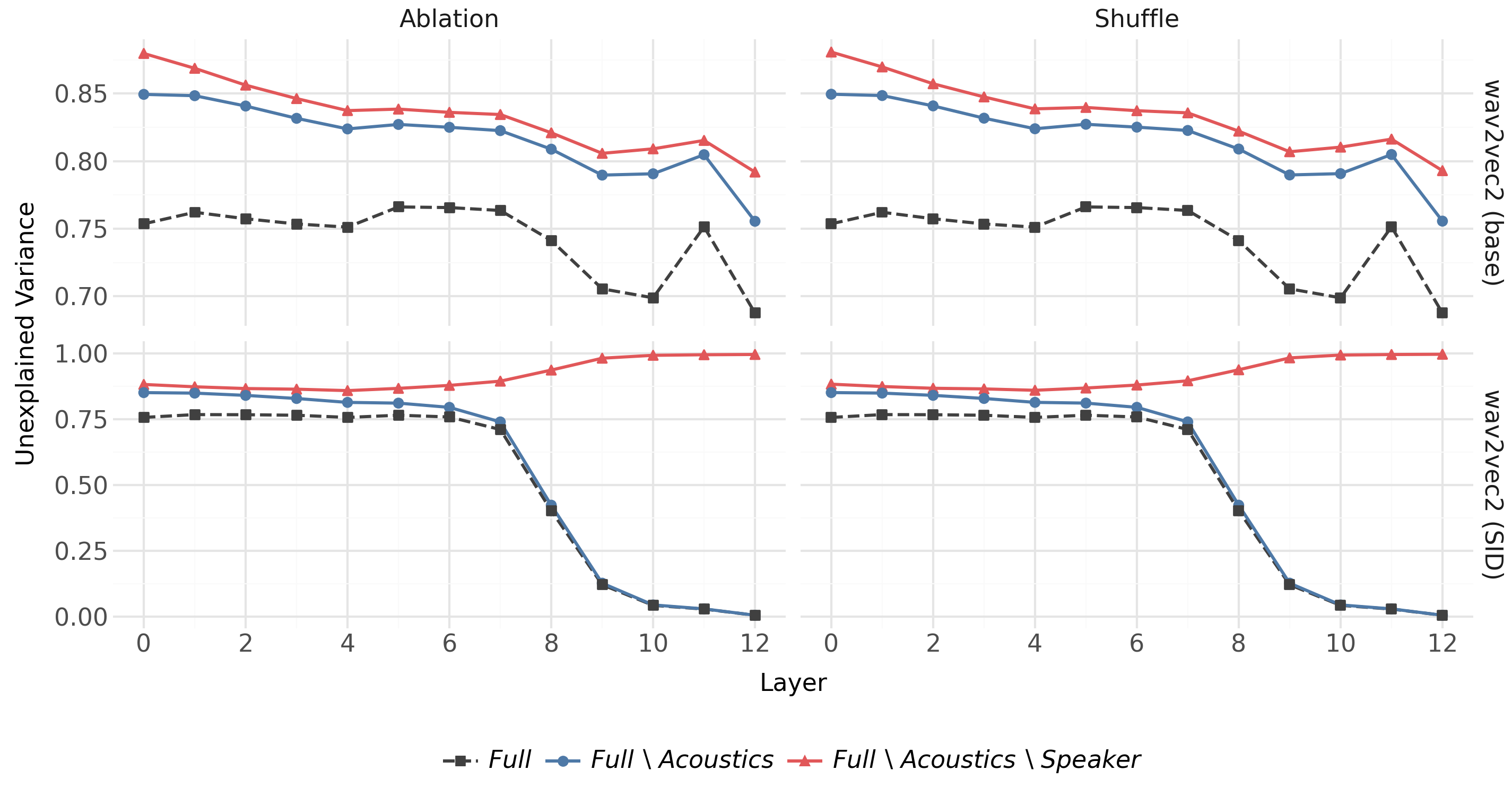}
  \caption{Comparison of the ablation (left column) and permutation (right
    column) controls for the speaker identity case study, for the base (top
    row) and speaker-identification (SID, bottom row) wav2vec2 models. Each
    panel reports the unexplained variance (UV) across layers, with color
    encoding the feature configuration. These runs use a single random seed.}
  \label{fig:control-comparison}
\end{figure}

\Cref{fig:control-comparison} compares the ablation (\textit{Ablation}, left
column) and permutation (\textit{Shuffle}, right column) variants for the base
(top row) and speaker-identification (SID, bottom row) models. For every layer,
model, and feature configuration, the permutation curves closely track their
ablation counterparts. This indicates that our \textsc{Encoding Probe} results
primarily reflect the effect of the feature removal rather than the
change in input dimensionality.

\subsection{PCA dimensionality-matched control}
\label{sec:pca-matched-control}
\begin{figure}
  \centering
  \includegraphics[width = \linewidth]{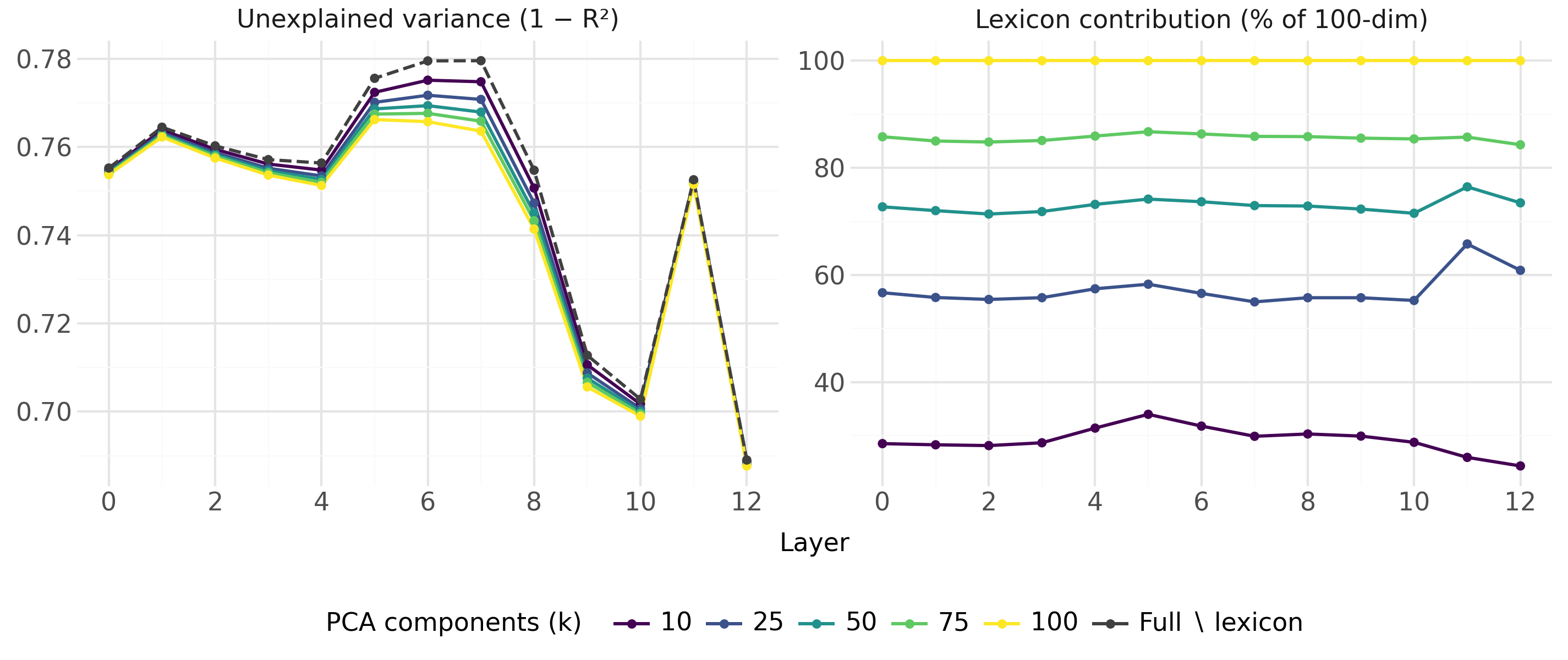}
  \caption{PCA dimensionality-matched control for the lexicon block on
    wav2vec2-base. \textit{Left:} unexplained variance (UV) against layer for
    the \textit{Full} probe with the lexicon block reduced to $k$ principal
    components (colored lines) and for the
    $\textit{Full}\setminus\textit{Lexicon}$ baseline (grey dashed line); the
    gap between the dashed baseline and each colored line is the lexicon
    contribution at that dimensionality. \textit{Right:} the same lexicon
    ablation gap at each $k$, expressed as a percentage of the gap obtained with
    the unreduced 100-dimensional representation, so the $k=100$ line is flat at
    $100\%$ and the remaining lines show the fraction of the full contribution
    retained at each reduced dimensionality.}
  \label{fig:lexicon-dim-sweep}
\end{figure}

The feature inventory we employ in the \textsc{Encoding Probe} analysis contains
feature blocks of differing dimensionality. It is reasonable to assume that a
richer, higher-dimensional feature block can account for more variance in the
representation space of our target models. Specifically in our experiment, the
lexicon block is a 100-dimensional fastText embedding, whereas the acoustic
block consists of 25 low-level descriptors and the syntactic block of a small
set of one-hot and scalar features. To isolate the effect of dimensionality, we
reduce the lexicon block to $k \in \{10, 25, 50, 75, 100\}$ principal components
and refit the encoding probe, measuring the lexicon ablation gap at each $k$.
\Cref{fig:lexicon-dim-sweep} shows the result for wav2vec2-base.

The lexicon contribution roughly scales as the number of dimensions increases.
The fastText embedding that is reduced to 25 dimensions (the dimensionality of
the acoustic block) with PCA retains on average about $57\%$ of its full
contribution, and at 10 dimensions about $29\%$ remains. A substantial part of
the lexicon block's measured absolute contribution therefore reflects its
dimensionality rather than lexical content alone. Hence we highlight the
cross-model comparisons on the same feature in the main text since these are the
more robust results from our \textsc{Encoding Probe} formulation.

\subsection{All model results}
\label{sec:all-model-results}
We also run the \textsc{Encoding Probe} analysis across an extended set of
speech and text Transformer-based models. We cover different architectures
(HuBERT, WavLM, RoBERTa, ModernBERT), sizes (12-layer, 24-layer), and training
objectives (speaker identification model from the SUPERB task). The complete
list of the extended model set is presented in \Cref{tab:extended-models}. The
speaker identification model from the main text is omitted to improve
readability of the results. Since the \textit{Full} probe of the SID model
reaches a near-zero UV score in the final layers and dominates the scale, we
omit the SID model's results here.

All extended-model runs use the same probing setup, feature sets, sampling
procedure, and train/test split described in \Cref{sec:method}, so differences
across models are attributable to representational differences rather than
pipeline changes. We extract hidden states layer-by-layer (including the
embedding/output layer, indexed as layer 0 in our plots), and apply identical
ablation-based encoding analyses as in the main experiments. Layer indices are
normalized to $[0, 1]$ for comparability across different model sizes. WavLM,
which uses a joint-speech enhancement objective, follows a broadly similar trend
to wav2vec2 and HuBERT base models.

\begin{table}
  \centering
  \begin{tabular}{ll}
    \toprule
    Model ID                          & Training     \\
    \midrule
    \texttt{wav2vec2-base}            & speech / PT  \\
    \texttt{wav2vec2-base-960h}       & speech / ASR \\
    \texttt{wav2vec2-base-superb-sid} & speech / SID \\
    \texttt{wav2vec2-large}           & speech / PT  \\
    \texttt{wav2vec2-large-960h}      & speech / ASR \\
    \texttt{hubert-base-ls960}        & speech / PT  \\
    \texttt{hubert-large-ll60k}       & speech / PT  \\
    \texttt{hubert-large-ls960-ft}    & speech / ASR \\
    \texttt{wavlm-base}               & speech / PT  \\
    \texttt{bert-base-uncased}        & text / PT    \\
    \texttt{roberta-base}             & text / PT    \\
    \texttt{ModernBERT-base}          & text / PT    \\
    \bottomrule
  \end{tabular}
  \caption{Extended model inventory used for additional probing runs. PT = Pre-training, ASR = Automatic Speech Recognition, SID = Speaker Identification.}
  \label{tab:extended-models}
\end{table}

In all figures below we report Unexplained Variance (UV = $1 - R^2$). The grey
dashed line shows the \textit{Full} probe without ablations; a larger upward gap
from this baseline means the ablated feature contributes more to reconstructing
the layer's representations. Only the gaps between curves---not the absolute UV
values---are the quantities of interest.

\subsubsection{Speaker Identity}
\begin{figure*}
  \centering
  \includegraphics[width=0.9\linewidth]{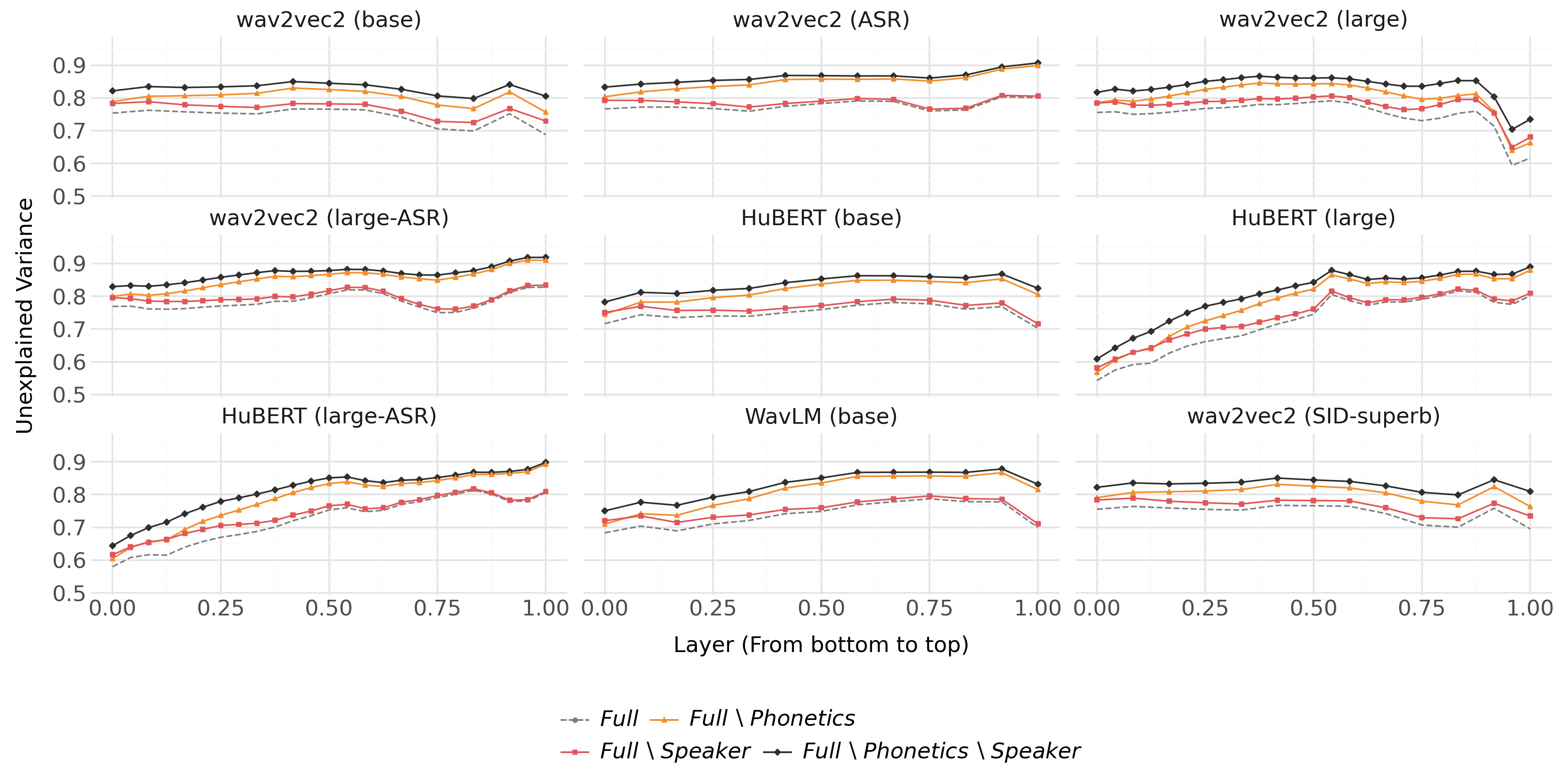}
  \caption{Encoding probe UV $(1-R^2)$ score results of phonetics and speaker
    identity for all models (larger gap from \textit{Full}/dashed = greater
    contribution).}
  \label{fig:all-model-phonetic-speaker}
\end{figure*}

\begin{figure*}
  \centering
  \includegraphics[width=0.9\linewidth]{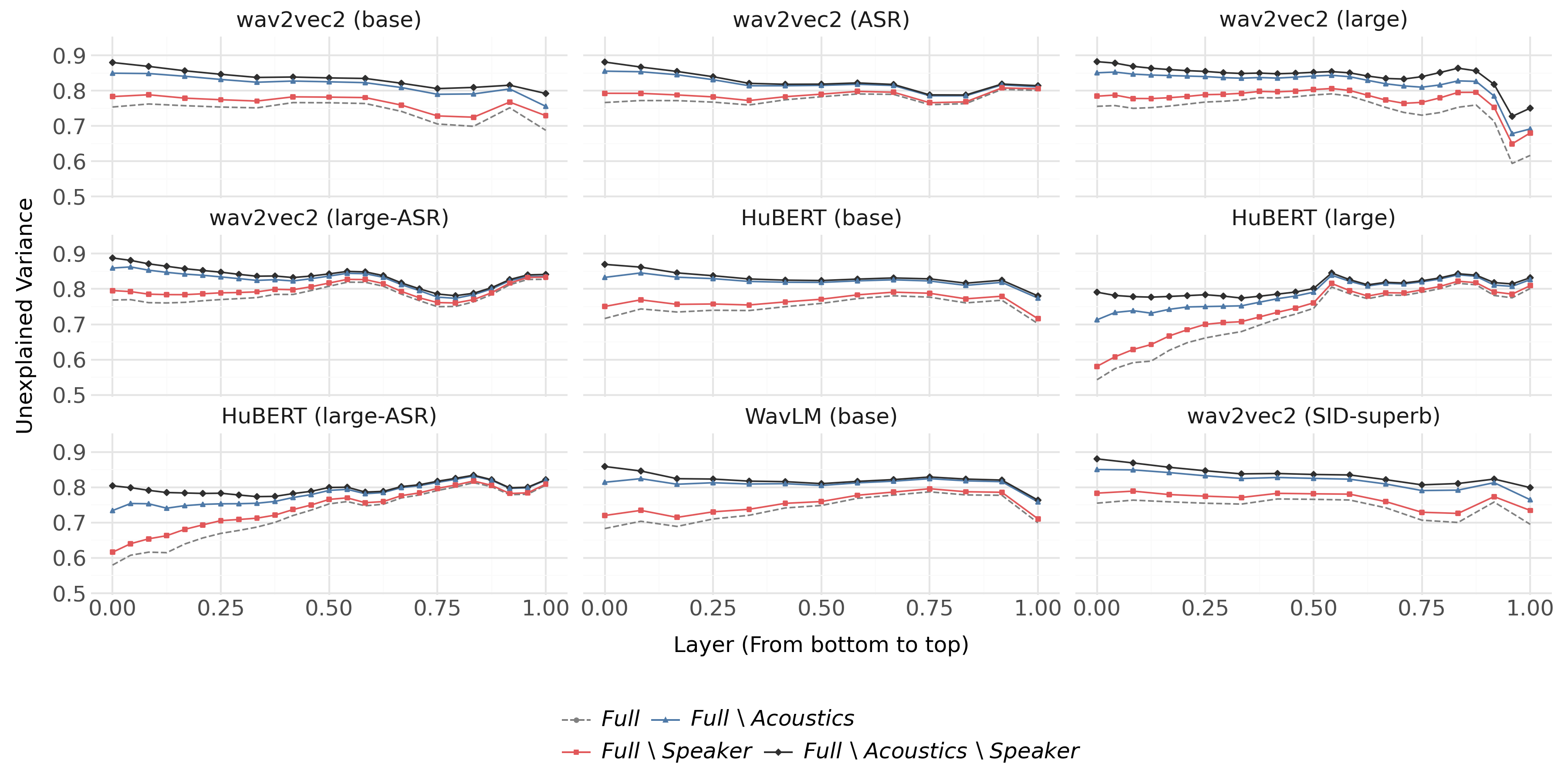}
  \caption{Encoding probe UV $(1-R^2)$ score results of acoustics and speaker
    identity for all models (larger gap from \textit{Full}/dashed = greater
    contribution).}
  \label{fig:all-model-acoustic-speaker}
\end{figure*}

We run the speaker identity experiment on the extended set of speech models
following the exact configuration in \Cref{sec:speaker-identity}. The results
are shown in
\Cref{fig:all-model-acoustic-speaker,fig:all-model-phonetic-speaker}. The
results from the base and large variants of wav2vec2 follow a very similar
trend, except in the final two layers where the large 24-layer model has a large
drop in Unexplained Variance. At the same time, the HuBERT base and large
variants mostly differ in the early layers where the gap between the
\textit{Full} probe and
$\textit{Full}\setminus\textit{Acoustics}\setminus\textit{Speaker}$ is much
larger for the 24-layer model than the 12-layer model. Our observation on the
ASR tuning objective in \Cref{sec:speaker-identity} holds across different model
architectures and sizes. Curiously, the SID-SUPERB model does not seem very
different from the base model it was based on. This could be due to two factors:
SID-superb is fine-tuned on the VoxCeleb dataset, which is out-of-distribution
for the current task, and the fine-tuning process may differ between SID-superb
and the SID model from the main text.

\subsubsection{Syntactic Information}
Turning to syntactic information, we again see a similar trend in different
model sizes: there are differences at the last three layers for wav2vec2 base
and large models and first six layers for HuBERT base and large models. We also
confirm that ASR-tuning increases the contribution from both lexical and
syntactic features in the large variants of both wav2vec2 and HuBERT. At the
same time, the probe results from the text models are quite stable and similar
across all three architectures.
\begin{figure*}
  \centering
  \includegraphics[width=0.9\linewidth]{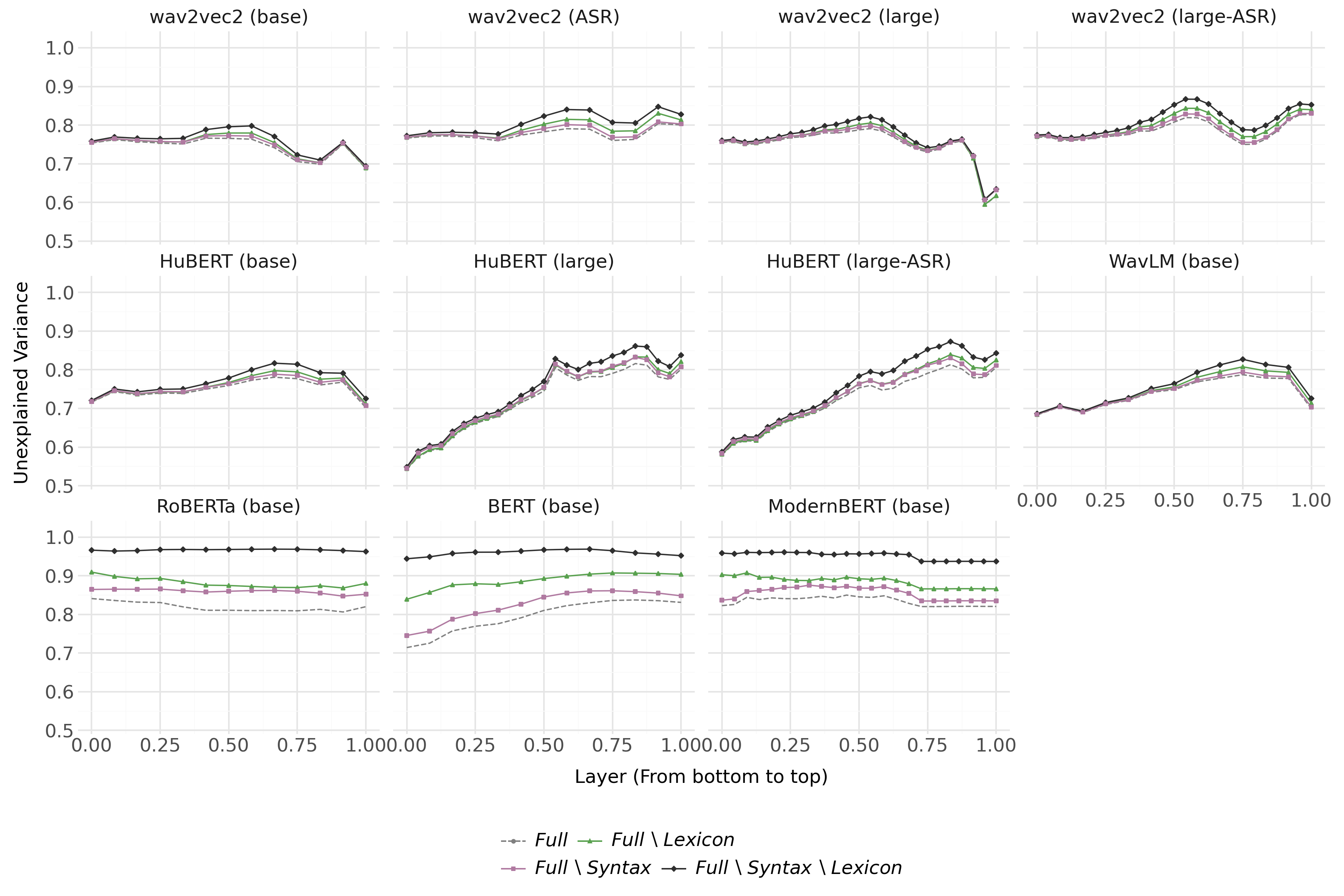}
  \caption{Encoding probe UV $(1-R^2)$ score results of syntax and lexicon for
    all models (larger gap from \textit{Full}/dashed = greater contribution).}
  \label{fig:all-model-syntax}
\end{figure*}


\begin{thebibliography}{41}
\providecommand{\natexlab}[1]{#1}

\bibitem[{Adi et~al.(2017)Adi, Kermany, Belinkov, Lavi, and
  Goldberg}]{adiFinegrainedAnalysisSentence2017}
Yossi Adi, Einat Kermany, Yonatan Belinkov, Ofer Lavi, and Yoav Goldberg. 2017.
\newblock Fine-grained {{Analysis}} of {{Sentence Embeddings Using Auxiliary
  Prediction Tasks}}.
\newblock In \emph{International {{Conference}} on {{Learning
  Representations}}}.

\bibitem[{Alain and Bengio(2017)}]{alainUnderstandingIntermediateLayers2017}
Guillaume Alain and Yoshua Bengio. 2017.
\newblock Understanding intermediate layers using linear classifier probes.
\newblock International Conference on Learning Representations.

\bibitem[{Baevski et~al.(2020)Baevski, Zhou, Mohamed, and
  Auli}]{baevskiWav2vec20Framework2020}
Alexei Baevski, Yuhao Zhou, Abdelrahman Mohamed, and Michael Auli. 2020.
\newblock Wav2vec 2.0: {{A Framework}} for {{Self-Supervised Learning}} of
  {{Speech Representations}}.
\newblock In \emph{Advances in {{Neural Information Processing Systems}}},
  volume~33, pages 12449--12460. Curran Associates, Inc.

\bibitem[{Belinkov(2022)}]{belinkovProbingClassifiersPromises2022}
Yonatan Belinkov. 2022.
\newblock \href {https://doi.org/10.1162/coli_a_00422} {Probing
  {{Classifiers}}: {{Promises}}, {{Shortcomings}}, and {{Advances}}}.
\newblock \emph{Computational Linguistics}, 48(1):207--219.

\bibitem[{Bentum et~al.(2025)Bentum, {ten Bosch}, and
  Lentz}]{bentumWordStressSelfsupervised2025}
Martijn Bentum, Louis {ten Bosch}, and Tomas~O. Lentz. 2025.
\newblock \href {https://doi.org/10.21437/Interspeech.2025-106} {Word stress in
  self-supervised speech models: {{A}} cross-linguistic comparison}.
\newblock In \emph{Proc. {{Interspeech}} 2025}, pages 251--255.

\bibitem[{Bojanowski et~al.(2017)Bojanowski, Grave, Joulin, and
  Mikolov}]{bojanowskiEnrichingWordVectors2017}
Piotr Bojanowski, Edouard Grave, Armand Joulin, and Tomas Mikolov. 2017.
\newblock \href {https://doi.org/10.1162/tacl_a_00051} {Enriching {{Word
  Vectors}} with {{Subword Information}}}.
\newblock \emph{Transactions of the Association for Computational Linguistics},
  5:135--146.

\bibitem[{Chiu et~al.(2026)Chiu, Fung, Li, Li, and
  Lee}]{chiuLargeScaleProbingAnalysis2026}
Aemon Yat~Fei Chiu, Kei~Ching Fung, Roger Tsz~Yeung Li, Jingyu Li, and Tan Lee.
  2026.
\newblock \href {https://doi.org/10.48550/arXiv.2501.05310} {A {{Large-Scale
  Probing Analysis}} of {{Speaker-Specific Attributes}} in {{Self-Supervised
  Speech Representations}}}.
\newblock \emph{Preprint}, arXiv:2501.05310.

\bibitem[{Choi et~al.(2024)Choi, Pasad, Nakamura, Fukayama, Livescu, and
  Watanabe}]{choiSelfSupervisedSpeechRepresentations2024}
Kwanghee Choi, Ankita Pasad, Tomohiko Nakamura, Satoru Fukayama, Karen Livescu,
  and Shinji Watanabe. 2024.
\newblock \href {https://doi.org/10.21437/Interspeech.2024-1157}
  {Self-{{Supervised Speech Representations}} are {{More Phonetic}} than
  {{Semantic}}}.
\newblock In \emph{Proc. {{Interspeech}} 2024}, pages 4578--4582.

\bibitem[{Chrupa{\l}a and
  Alishahi(2019)}]{chrupalaCorrelatingNeuralSymbolic2019}
Grzegorz Chrupa{\l}a and Afra Alishahi. 2019.
\newblock \href {https://doi.org/10.18653/v1/P19-1283} {Correlating {{Neural}}
  and {{Symbolic Representations}} of {{Language}}}.
\newblock In \emph{Proceedings of the 57th {{Annual Meeting}} of the
  {{Association}} for {{Computational Linguistics}}}, pages 2952--2962,
  Florence, Italy. Association for Computational Linguistics.

\bibitem[{Churchwell et~al.(2024)Churchwell, Morrison, and
  Pardo}]{churchwellHighFidelityNeuralPhonetic2024}
Cameron Churchwell, Max Morrison, and Bryan Pardo. 2024.
\newblock \href {https://doi.org/10.1109/ICASSPW62465.2024.10669905}
  {High-{{Fidelity Neural Phonetic Posteriorgrams}}}.
\newblock In \emph{2024 {{IEEE International Conference}} on {{Acoustics}},
  {{Speech}}, and {{Signal Processing Workshops}} ({{ICASSPW}})}, pages
  823--827, Seoul, Korea, Republic of. IEEE.

\bibitem[{Conneau et~al.(2018)Conneau, Kruszewski, Lample, Barrault, and
  Baroni}]{conneauWhatYouCan2018}
Alexis Conneau, German Kruszewski, Guillaume Lample, Lo{\"i}c Barrault, and
  Marco Baroni. 2018.
\newblock \href {https://doi.org/10.18653/v1/P18-1198} {What you can cram into
  a single \$\&!\#* vector: {{Probing}} sentence embeddings for linguistic
  properties}.
\newblock In \emph{Proceedings of the 56th {{Annual Meeting}} of the
  {{Association}} for {{Computational Linguistics}} ({{Volume}} 1: {{Long
  Papers}})}, pages 2126--2136, Melbourne, Australia. Association for
  Computational Linguistics.

\bibitem[{Cormac~English et~al.(2022)Cormac~English, Kelleher, and
  {Carson-Berndsen}}]{cormacenglishDomainInformedProbingWav2vec2022}
Patrick Cormac~English, John~D. Kelleher, and Julie {Carson-Berndsen}. 2022.
\newblock \href {https://doi.org/10.18653/v1/2022.sigmorphon-1.9}
  {Domain-{{Informed Probing}} of wav2vec 2.0 {{Embeddings}} for {{Phonetic
  Features}}}.
\newblock In \emph{Proceedings of the 19th {{SIGMORPHON Workshop}} on
  {{Computational Research}} in {{Phonetics}}, {{Phonology}}, and
  {{Morphology}}}, pages 83--91, Seattle, Washington. Association for
  Computational Linguistics.

\bibitem[{{de Heer Kloots} et~al.(2025){de Heer Kloots}, Mohebbi, Pouw, Shen,
  Zuidema, and Bentum}]{deheerklootsWhatSelfsupervisedSpeech2025}
Marianne {de Heer Kloots}, Hosein Mohebbi, Charlotte Pouw, Gaofei Shen, Willem
  Zuidema, and Martijn Bentum. 2025.
\newblock \href {https://doi.org/10.21437/Interspeech.2025-1526} {What do
  self-supervised speech models know about {{Dutch}}? {{Analyzing}} advantages
  of language-specific pre-training}.
\newblock In \emph{Proc. {{Interspeech}} 2025}, pages 256--260.

\bibitem[{De~La~Fuente and
  Jurafsky(2024)}]{delafuenteLayerwiseAnalysisMandarin2024}
Anton De~La~Fuente and Dan Jurafsky. 2024.
\newblock \href {https://doi.org/10.21437/Interspeech.2024-2341} {A layer-wise
  analysis of {{Mandarin}} and {{English}} suprasegmentals in {{SSL}} speech
  models}.
\newblock In \emph{Interspeech 2024}, pages 1290--1294. ISCA.

\bibitem[{Devlin et~al.(2019)Devlin, Chang, Lee, and
  Toutanova}]{devlin-etal-2019-bert}
Jacob Devlin, Ming-Wei Chang, Kenton Lee, and Kristina Toutanova. 2019.
\newblock \href {https://doi.org/10.18653/v1/N19-1423} {{{BERT}}:
  {{Pre-training}} of deep bidirectional transformers for language
  understanding}.
\newblock In \emph{Proceedings of the 2019 Conference of the North {{American}}
  Chapter of the Association for Computational Linguistics: {{Human}} Language
  Technologies, Volume 1 (Long and Short Papers)}, pages 4171--4186,
  Minneapolis, Minnesota. Association for Computational Linguistics.

\bibitem[{Eyben et~al.(2016)Eyben, Scherer, Schuller, Sundberg, Andre, Busso,
  Devillers, Epps, Laukka, Narayanan, and
  Truong}]{eybenGenevaMinimalisticAcoustic2016}
Florian Eyben, Klaus~R. Scherer, Bjorn~W. Schuller, Johan Sundberg, Elisabeth
  Andre, Carlos Busso, Laurence~Y. Devillers, Julien Epps, Petri Laukka,
  Shrikanth~S. Narayanan, and Khiet~P. Truong. 2016.
\newblock \href {https://doi.org/10.1109/TAFFC.2015.2457417} {The {{Geneva
  Minimalistic Acoustic Parameter Set}} ({{GeMAPS}}) for {{Voice Research}} and
  {{Affective Computing}}}.
\newblock \emph{IEEE Transactions on Affective Computing}, 7(2):190--202.

\bibitem[{Eyben et~al.(2010)Eyben, W{\"o}llmer, and
  Schuller}]{eybenOpensmileMunichVersatile2010}
Florian Eyben, Martin W{\"o}llmer, and Bj{\"o}rn Schuller. 2010.
\newblock \href {https://doi.org/10.1145/1873951.1874246} {Opensmile: The
  munich versatile and fast open-source audio feature extractor}.
\newblock In \emph{Proceedings of the 18th {{ACM}} International Conference on
  {{Multimedia}}}, pages 1459--1462, Firenze Italy. ACM.

\bibitem[{Gubian et~al.(2025)Gubian, Krehan, Liu, Kirby, and
  Goldwater}]{gubianAnalyzingRelationshipsPretraining2025}
Michele Gubian, Ioana Krehan, Oli Liu, James Kirby, and Sharon Goldwater. 2025.
\newblock \href {https://doi.org/10.48550/arXiv.2506.10855} {Analyzing the
  relationships between pretraining language, phonetic, tonal, and speaker
  information in self-supervised speech models}.
\newblock \emph{Preprint}, arXiv:2506.10855.

\bibitem[{Hazen et~al.(2009)Hazen, Shen, and
  White}]{hazenQuerybyexampleSpokenTerm2009}
Timothy~J. Hazen, Wade Shen, and Christopher White. 2009.
\newblock \href {https://doi.org/10.1109/ASRU.2009.5372889} {Query-by-example
  spoken term detection using phonetic posteriorgram templates}.
\newblock In \emph{2009 {{IEEE Workshop}} on {{Automatic Speech Recognition}}
  \& {{Understanding}}}, pages 421--426.

\bibitem[{Hewitt et~al.(2021)Hewitt, Ethayarajh, Liang, and
  Manning}]{hewittConditionalProbingMeasuring2021}
John Hewitt, Kawin Ethayarajh, Percy Liang, and Christopher Manning. 2021.
\newblock \href {https://doi.org/10.18653/v1/2021.emnlp-main.122} {Conditional
  probing: Measuring usable information beyond a baseline}.
\newblock In \emph{Proceedings of the 2021 {{Conference}} on {{Empirical
  Methods}} in {{Natural Language Processing}}}, pages 1626--1639, Online and
  Punta Cana, Dominican Republic. Association for Computational Linguistics.

\bibitem[{Hewitt and Liang(2019)}]{hewittDesigningInterpretingProbes2019}
John Hewitt and Percy Liang. 2019.
\newblock \href {https://doi.org/10.18653/v1/D19-1275} {Designing and
  {{Interpreting Probes}} with {{Control Tasks}}}.
\newblock In \emph{Proceedings of the 2019 {{Conference}} on {{Empirical
  Methods}} in {{Natural Language Processing}} and the 9th {{International
  Joint Conference}} on {{Natural Language Processing}} ({{EMNLP-IJCNLP}})},
  pages 2733--2743, Hong Kong, China. Association for Computational
  Linguistics.

\bibitem[{Hewitt and Manning(2019)}]{hewittStructuralProbeFinding2019}
John Hewitt and Christopher~D. Manning. 2019.
\newblock \href {https://doi.org/10.18653/v1/N19-1419} {A {{Structural Probe}}
  for {{Finding Syntax}} in {{Word Representations}}}.
\newblock In \emph{Proceedings of the 2019 {{Conference}} of the {{North
  American Chapter}} of the {{Association}} for {{Computational Linguistics}}:
  {{Human Language Technologies}}, {{Volume}} 1 ({{Long}} and {{Short
  Papers}})}, pages 4129--4138, Minneapolis, Minnesota. Association for
  Computational Linguistics.

\bibitem[{Honnibal et~al.(2020)Honnibal, Montani, Van~Landeghem, and
  Boyd}]{honnibalSpaCyIndustrialstrengthNatural2020}
Matthew Honnibal, Ines Montani, Sofie Van~Landeghem, and Adriane Boyd. 2020.
\newblock \href {https://doi.org/10.5281/zenodo.1212303} {{{spaCy}}:
  {{Industrial-strength}} natural language processing in python}.

\bibitem[{Hotelling(1936)}]{hotellingRelationsTwoSets1936}
Harold Hotelling. 1936.
\newblock \href {https://doi.org/10.2307/2333955} {Relations {{Between Two
  Sets}} of {{Variates}}}.
\newblock \emph{Biometrika}, 28(3/4):321--377.

\bibitem[{Hsu et~al.(2021)Hsu, Bolte, Tsai, Lakhotia, Salakhutdinov, and
  Mohamed}]{hsuHuBERTSelfSupervisedSpeech2021}
Wei-Ning Hsu, Benjamin Bolte, Yao-Hung~Hubert Tsai, Kushal Lakhotia, Ruslan
  Salakhutdinov, and Abdelrahman Mohamed. 2021.
\newblock \href {https://doi.org/10.1109/TASLP.2021.3122291} {{{HuBERT}}:
  {{Self-Supervised Speech Representation Learning}} by {{Masked Prediction}}
  of {{Hidden Units}}}.
\newblock \emph{IEEE/ACM Transactions on Audio, Speech, and Language
  Processing}, 29:3451--3460.

\bibitem[{Hupkes and
  Zuidema(2018)}]{hupkesVisualisationDiagnosticClassifiers2018}
Dieuwke Hupkes and Willem Zuidema. 2018.
\newblock Visualisation and '{{Diagnostic Classifiers}}' {{Reveal}} how
  {{Recurrent}} and {{Recursive Neural Networks Process Hierarchical
  Structure}} ({{Extended Abstract}}).
\newblock pages 5617--5621.

\bibitem[{Ivanova et~al.(2021)Ivanova, Hewitt, and
  Zaslavsky}]{ivanova2021probing}
Anna~A Ivanova, John Hewitt, and Noga Zaslavsky. 2021.
\newblock Probing artificial neural networks: {{Insights}} from neuroscience.
\newblock In \emph{{{ICLR}} 2021 Workshop ``How Can Findings about the Brain
  Improve {{AI}} Systems?''}.

\bibitem[{Kisler et~al.(2017)Kisler, Reichel, and
  Schiel}]{kislerMultilingualProcessingSpeech2017}
Thomas Kisler, Uwe Reichel, and Florian Schiel. 2017.
\newblock \href {https://doi.org/10.1016/j.csl.2017.01.005} {Multilingual
  processing of speech via web services}.
\newblock \emph{Computer Speech \& Language}, 45:326--347.

\bibitem[{Kriegeskorte et~al.(2008)Kriegeskorte, Mur, and
  Bandettini}]{kriegeskorteRepresentationalSimilarityAnalysisconnecting2008}
Nikolaus Kriegeskorte, Marieke Mur, and Peter~A Bandettini. 2008.
\newblock Representational similarity analysis-connecting the branches of
  systems neuroscience.
\newblock \emph{Frontiers in systems neuroscience}, page~4.

\bibitem[{Lee et~al.(2019)Lee, Keating, and
  Kreiman}]{leeAcousticVoiceVariation2019}
Yoonjeong Lee, Patricia Keating, and Jody Kreiman. 2019.
\newblock \href {https://doi.org/10.1121/1.5125134} {Acoustic voice variation
  within and between speakers}.
\newblock \emph{The Journal of the Acoustical Society of America},
  146(3):1568--1579.

\bibitem[{Liu et~al.(2023)Liu, Tang, and
  Goldwater}]{liuSelfsupervisedPredictiveCoding2023}
Oli~Danyi Liu, Hao Tang, and Sharon Goldwater. 2023.
\newblock \href {https://doi.org/10.21437/Interspeech.2023-871}
  {Self-supervised {{Predictive Coding Models Encode Speaker}} and {{Phonetic
  Information}} in {{Orthogonal Subspaces}}}.
\newblock In \emph{Proc. {{Interspeech}} 2023}, pages 2968--2972.

\bibitem[{Ma et~al.(2021)Ma, Ryant, and
  Liberman}]{maProbingAcousticRepresentations2021}
Danni Ma, Neville Ryant, and Mark Liberman. 2021.
\newblock \href {https://doi.org/10.1109/ICASSP39728.2021.9414776} {Probing
  {{Acoustic Representations}} for {{Phonetic Properties}}}.
\newblock In \emph{{{ICASSP}} 2021 - 2021 {{IEEE International Conference}} on
  {{Acoustics}}, {{Speech}} and {{Signal Processing}} ({{ICASSP}})}, pages
  311--315.

\bibitem[{Maudslay and Cotterell(2021)}]{maudslaySyntacticProbesProbe2021}
Rowan~Hall Maudslay and Ryan Cotterell. 2021.
\newblock \href {https://doi.org/10.18653/v1/2021.naacl-main.11} {Do
  {{Syntactic Probes Probe Syntax}}? {{Experiments}} with {{Jabberwocky
  Probing}}}.
\newblock In \emph{Proceedings of the 2021 {{Conference}} of the {{North
  American Chapter}} of the {{Association}} for {{Computational Linguistics}}:
  {{Human Language Technologies}}}, pages 124--131, Online. Association for
  Computational Linguistics.

\bibitem[{Panayotov et~al.(2015)Panayotov, Chen, Povey, and
  Khudanpur}]{panayotovLibrispeechASRCorpus2015}
Vassil Panayotov, Guoguo Chen, Daniel Povey, and Sanjeev Khudanpur. 2015.
\newblock \href {https://doi.org/10.1109/ICASSP.2015.7178964} {Librispeech:
  {{An ASR}} corpus based on public domain audio books}.
\newblock In \emph{2015 {{IEEE International Conference}} on {{Acoustics}},
  {{Speech}} and {{Signal Processing}} ({{ICASSP}})}, pages 5206--5210.

\bibitem[{Pasad et~al.(2024)Pasad, Chien, Settle, and
  Livescu}]{pasadWhatSelfSupervisedSpeech2024}
Ankita Pasad, Chung-Ming Chien, Shane Settle, and Karen Livescu. 2024.
\newblock \href {https://doi.org/10.1162/tacl_a_00656} {What {{Do
  Self-Supervised Speech Models Know About Words}}?}
\newblock \emph{Transactions of the Association for Computational Linguistics},
  12:372--391.

\bibitem[{Pasad et~al.(2021)Pasad, Chou, and
  Livescu}]{pasadLayerwiseAnalysisSelfsupervised2021}
Ankita Pasad, Ju-Chieh Chou, and Karen Livescu. 2021.
\newblock \href {https://doi.org/10.1109/ASRU51503.2021.9688093} {Layer-{{Wise
  Analysis}} of a {{Self-Supervised Speech Representation Model}}}.
\newblock In \emph{2021 {{IEEE Automatic Speech Recognition}} and
  {{Understanding Workshop}} ({{ASRU}})}, pages 914--921, Cartagena, Colombia.
  IEEE.

\bibitem[{Pedregosa et~al.(2011)Pedregosa, Varoquaux, Gramfort, Michel,
  Thirion, Grisel, Blondel, Prettenhofer, Weiss, Dubourg, Vanderplas, Passos,
  Cournapeau, Brucher, Perrot, and
  Duchesnay}]{pedregosaScikitlearnMachineLearning2011}
F.~Pedregosa, G.~Varoquaux, A.~Gramfort, V.~Michel, B.~Thirion, O.~Grisel,
  M.~Blondel, P.~Prettenhofer, R.~Weiss, V.~Dubourg, J.~Vanderplas, A.~Passos,
  D.~Cournapeau, M.~Brucher, M.~Perrot, and E.~Duchesnay. 2011.
\newblock Scikit-learn: {{Machine}} learning in {{Python}}.
\newblock \emph{Journal of Machine Learning Research}, 12:2825--2830.

\bibitem[{Shen et~al.(2023)Shen, Alishahi, Bisazza, and
  Chrupa{\l}a}]{shenWaveSyntaxProbing2023}
Gaofei Shen, Afra Alishahi, Arianna Bisazza, and Grzegorz Chrupa{\l}a. 2023.
\newblock \href {https://doi.org/10.21437/Interspeech.2023-679} {Wave to
  {{Syntax}}: {{Probing}} spoken language models for syntax}.
\newblock In \emph{{{INTERSPEECH}} 2023}, pages 1259--1263.

\bibitem[{Simon et~al.(2025)Simon, Chemla, King, and
  Lakretz}]{simon2025probing}
Pablo J.~Diego Simon, Emmanuel Chemla, Jean-Remi King, and Yair Lakretz. 2025.
\newblock Probing syntax in large language models: {{Successes}} and remaining
  challenges.
\newblock In \emph{Second Conference on Language Modeling}.

\bibitem[{Tenney et~al.(2019)Tenney, Das, and
  Pavlick}]{tenneyBERTRediscoversClassical2019}
Ian Tenney, Dipanjan Das, and Ellie Pavlick. 2019.
\newblock \href {https://doi.org/10.18653/v1/P19-1452} {{{BERT Rediscovers}}
  the {{Classical NLP Pipeline}}}.
\newblock In \emph{Proceedings of the 57th {{Annual Meeting}} of the
  {{Association}} for {{Computational Linguistics}}}, pages 4593--4601,
  Florence, Italy. Association for Computational Linguistics.

\bibitem[{Wolf et~al.(2020)Wolf, Debut, Sanh, Chaumond, Delangue, Moi, Cistac,
  Rault, Louf, Funtowicz, Davison, Shleifer, {von Platen}, Ma, Jernite, Plu,
  Xu, Le~Scao, Gugger, Drame, Lhoest, and
  Rush}]{wolfTransformersStateoftheArtNatural2020}
Thomas Wolf, Lysandre Debut, Victor Sanh, Julien Chaumond, Clement Delangue,
  Anthony Moi, Pierric Cistac, Tim Rault, Remi Louf, Morgan Funtowicz, Joe
  Davison, Sam Shleifer, Patrick {von Platen}, Clara Ma, Yacine Jernite, Julien
  Plu, Canwen Xu, Teven Le~Scao, Sylvain Gugger, and 3 others. 2020.
\newblock \href {https://doi.org/10.18653/v1/2020.emnlp-demos.6} {Transformers:
  {{State-of-the-Art Natural Language Processing}}}.
\newblock In \emph{Proceedings of the 2020 {{Conference}} on {{Empirical
  Methods}} in {{Natural Language Processing}}: {{System Demonstrations}}},
  pages 38--45, Online. Association for Computational Linguistics.

\end{thebibliography}
\end{document}